\documentclass[final,times,5p,twocolumn]{elsarticle}
\usepackage{lineno}
\usepackage{amssymb}
\usepackage{amsmath}
\usepackage[ruled,linesnumbered]{algorithm2e}
\usepackage{xcolor}
\usepackage{mdframed}
\usepackage[ruled,linesnumbered]{algorithm2e}
\usepackage{amsmath,amssymb,amsfonts}
\usepackage{graphicx}
\usepackage{epstopdf}
\usepackage{subfig}
\usepackage{gensymb}
\usepackage{subfig}
\usepackage{multirow,booktabs,color,soul,threeparttable}
\usepackage{xcolor}
\usepackage{mdframed}
\usepackage{cleveref}
\usepackage{rotating}
\usepackage{booktabs}
\usepackage{array} 
\usepackage{float}

\newcommand{\RNum}[1]{\uppercase\expandafter{\romannumeral #1\relax}}
\definecolor{hl}{rgb}{0.75,0.75,0.75}
\sethlcolor{hl}
\definecolor{emph}{rgb}{0,0,1}

\begin{document}
\begin{frontmatter}
\title{Un-evaluated Solutions May Be Valuable in Expensive Optimization}

\author[sjtu]{Hao Hao} \ead{haohao@sjtu.edu.cn}
\author[sjtu]{Xiaoqun Zhang} \ead{xqzhang@sjtu.edu.cn}
\author[ecnu,aiedu]{Aimin Zhou\corref{mycorrespondingauthor}} \ead{amzhou@cs.ecnu.edu.cn}\cortext[mycorrespondingauthor]{Corresponding author: Aimin Zhou}

%% Author affiliation
\affiliation[sjtu]{organization={Institute of Natural Sciences},%Department and Organization
            addressline={Shanghai Jiao Tong University}, 
            city={Shanghai},
            postcode={200240}, 
          %   state={Shanghai},
            country={China}}

\affiliation[ecnu]{organization={School of Computer Science and Technology},%Department and Organization
addressline={East China Normal University}, 
city={Shanghai},
postcode={200062}, 
%   state={Shanghai},
country={China}}

\affiliation[aiedu]{organization={Shanghai Institute of AI for Education},%Department and Organization
addressline={East China Normal University}, 
city={Shanghai},
postcode={200062}, 
%   state={Shanghai},
country={China}}

\begin{abstract}
Expensive optimization problems~(EOPs) are prevalent in real-world applications, where the evaluation of a single solution requires a significant amount of resources. In our study of surrogate-assisted evolutionary algorithms (SAEAs) in EOPs, we discovered an intriguing phenomenon. Because only a limited number of solutions are evaluated in each iteration, relying solely on these evaluated solutions for evolution can lead to reduced disparity in successive populations. This, in turn, hampers the reproduction operators' ability to generate superior solutions, thereby reducing the algorithm's convergence speed. To address this issue, we propose a strategic approach that incorporates high-quality, un-evaluated solutions predicted by surrogate models during the selection phase. This approach aims to improve the distribution of evaluated solutions, thereby generating a superior next generation of solutions. This work details specific implementations of this concept across various reproduction operators and validates its effectiveness using multiple surrogate models. Experimental results demonstrate that the proposed strategy significantly enhances the performance of surrogate-assisted evolutionary algorithms. Compared to mainstream SAEAs and Bayesian optimization algorithms, our approach incorporating the un-evaluated solution strategy shows a marked improvement.
\end{abstract}


\end{frontmatter}

\section{Introduction}
Expensive optimization problems (EOPs) are common in various real-world scenarios where assessing a single solution requires a considerable amount of computational resources~\cite{li2022evolutionary,DBLP:journals/fcsc/WuQLZZ23}. Evolutionary algorithms (EAs) have gained widespread acceptance for solving a diverse array of problems across different domains, thanks to their ability to perform global searches and adapt to specific problem characteristics \cite{back1997handbook}. The assumption in most EAs is that the fitness of all solutions within a population can be evaluated. Typically, this is done using an explicit fitness function, computational simulations \cite{li2020surrogateassisted}, or through direct experimentation \cite{chugh2017datadrivena}. Nevertheless, practical challenges arise when fitness evaluations are costly, as seen in EOPs where each assessment demands a resource-intensive simulation or experimental procedure \cite{jin2011surrogateassisted}. Under these conditions, conventional EAs become inefficient, requiring a large number of evaluations to identify an optimal solution. Surrogate-assisted evolutionary algorithms (SAEAs) have been developed to mitigate these inefficiencies. In SAEAs, surrogate models are employed to approximate the true fitness landscape, thereby minimizing the need for expensive solution evaluations. These models serve as estimators for the actual fitness function, enabling the algorithm to efficiently explore the search space. By leveraging surrogate models, SAEAs retain the comprehensive search abilities of traditional EAs while significantly speeding up the optimization process. This makes SAEAs particularly well-suited for tackling intricate real-world problems that have limited computational budgets.

The fundamental paradigm of SAEAs involves utilizing surrogate models to predict solution quality before applying the selection operator within the evolutionary algorithm (EA) framework. By using these computational cheap models to identify promising solutions for actual evaluation, the number of costly evaluations per iteration can be significantly reduced, thereby enhancing the algorithm's efficiency. Based on the number $\gamma$ of solutions selected for real evaluations with the assistance of the surrogate model, SAEAs can be categorized into three types:

\begin{itemize}
\item $\gamma=1$: In each generation, only the best predicted solution is selected for real function evaluation and preserved in an archive. This strategy notably decreases the real evaluation budget. However, the downside is that only one solution in the next-generation population will be modified, which might limit the enhancement of population distribution and keep new solutions confined to the current search space. To tackle this issue, some studies have utilized acquisition functions to promote solution diversity. For instance, GPEME \cite{liu2014gaussian} and SADE-Sammon \cite{chen2020surrogate} leverage the lower confidence bound (LCB) \cite{srinivas2009gaussian} to guide the search. Similarly, efficient global optimization~(EGO)~\cite{jones1998efficient} employs the expected improvement~(EI) method~\cite{mockus1998application}, and SA-EDA \cite{hao2022surrogate} combines multiple acquisition strategies using the GP-Hedge method~\cite{DBLP:conf/uai/HoffmanBF11} to enhance the robustness of the selection process.

\item $\gamma=N$: Some methods use surrogate-assisted selection as a pre-selection step, generating more trial solutions than the population size $N$ during the reproduction phase. The model then predicts the outcomes, selecting the approximately best $N$ solutions for real evaluation. Although this approach effectively improves the quality of each generation's offspring, it does not significantly reduce the number of real evaluations. Examples include methods based on binary and fuzzy classification \cite{zhang2018preselection,zhou2019fuzzyclassification}, approaches using relation models for selection \cite{hao2020binary}, and techniques that introduce some solutions directly into the next generation without model evaluation \cite{LiExpensiveOptimizationSurrogateAssisted2022a}.

\item $1<\gamma<N$: Customized approaches have been proposed to balance the number of real evaluations and solution diversity. For instance, SAMSO \cite{li2020surrogateassisted} and SACOSO \cite{sun2017surrogateassisted} employ multi-particle swarm optimization to enhance diversity through interactions among swarms. Meanwhile, LLSO \cite{wei2020classifier} and DFC-MOEA \cite{zhang2022dual} use hierarchical strategies for solution selection. LLSO increases population diversity by introducing random solutions, whereas DFC-MOEA selects solutions with medium membership degrees using a classifier.
\end{itemize}

Through the carefully designed model-assisted selection strategies introduced above, the evaluation overhead can be effectively reduced. When dealing with EOPs, it is preferable to choose a smaller $\gamma$ to enable more iterations and explorations with a limited number of fitness evaluations (FEs). However, an important phenomenon to note is that during the algorithmic iterations, only a limited number of solutions are evaluated. Relying solely on these evaluated solutions to generate new offspring often fails to significantly improve the distribution of new solutions. This limitation can adversely affect the ability of reproduction operators to produce high-quality solutions, thereby reducing the algorithm's convergence speed. The work presented in \cite{hao2024enhancing} demonstrates a phenomenon akin to ``standing still'' during the algorithm's evolution process via low-dimensional visualization. This issue is prevalent in SAEAs, yet it is often unaddressed by existing reproduction operators, thus limiting the practical performance of SAEAs. Even the most accurate surrogate models cannot select high-quality individuals from low-quality evaluated solutions alone.

To address this problem, we propose a general conceptual strategy. During the algorithm iteration, high-quality un-evaluated solutions~($\mathcal{P}_u$) predicted by the surrogate models are combined with the evaluated solutions~($\mathcal{P}_e$) to form the next generation. These un-evaluated solutions, although not directly selected for real evaluations due to their slightly lower predicted quality, still have the potential to enhance the overall population and should not be discarded outright. This strategy aims to moderately increase the population's diversity compared to the previous generation, improve the effectiveness of reproduction operators, and thereby accelerate the optimization process.

To the best of our knowledge, Jin~\cite{jinSurrogateassistedEvolutionaryComputation2011a} in their review summarized a method known as the $(\mu+\lambda)$-Best strategy, which involves evaluating fewer than $\lambda$ individuals, while substituting the others with predictions from surrogate models. This method bears some similarity to our work. However, it emphasizes a specific model management strategy, whereas our work investigates the impact of unevaluated solutions on the quality of the offspring. Moreover, we implement this approach on three typical reproduction operators: genetic algorithm (GA)~\cite{deb1996combined,deb2007self} operators, differential evolution (DE)~\cite{das2010differential} operators, and estimation of distribution algorithm (EDA)~\cite{zhouEstimationDistributionAlgorithm2015} operators. This method is intended to improve the diversity of the population and the performance of the reproduction operators. We demonstrate the effectiveness of this strategy across various reproduction operators and surrogate models, highlighting its significant enhancement of SAEAs' performance. Experimental results show that our proposed strategy outperforms mainstream SAEAs and Bayesian optimization algorithms, indicating its effectiveness in accelerating the optimization process.

The main contributions of this work are as follows:

\begin{itemize}
    \item We propose a general strategy that combines high-quality un-evaluated solutions predicted by the surrogate models with evaluated solutions during the selection process to generate the next generation. This method aims to increase the population's disparity compared to the previous generation, enhance the performance of reproduction operators, and accelerate the optimization process.
    \item As a general conceptual strategy, we provide specific implementations on three types of reproduction operators: GA operators, DE operators, and EDA operators. Three surrogate models are embedded within the proposed strategy.
    \item We thoroughly verify the effectiveness of the proposed strategy through extensive experiments. The results show that the proposed strategy significantly improves the performance of SAEAs. Compared to mainstream SAEAs and Bayesian optimizations~(BOs), the proposed strategy demonstrates superior performance and attempts to explain the advantages of this strategy in overcoming model uncertainty compared to BOs.
\end{itemize}

The subsequent sections of this paper are organized as follows. Section~\ref{sec:preliminaries} introduces the preliminaries, including problem formulation, surrogate models, model-assisted strategies, and reproduction operators. Section~\ref{sec:motivation} presents the motivation for the proposed strategy. Section~\ref{sec:proposed_alg} details the proposed algorithm, encompassing the framework using un-evaluated solutions and the specific implementation of operators. Section~\ref{sec:experiments} presents the experimental results. Finally, Section~\ref{sec:conclusion} concludes the paper.

\section{Preliminaries}\label{sec:preliminaries}

\subsection{Problem Formulation}

An unconstrained minimization expensive optimization problem can be described as follows:
\begin{align}
 \label{equ:func}
 \min_{\mathbf{x} \in \Omega}~f(\mathbf{x}) 
 \end{align}
Here, $\mathbf{x}=(x_1,\ldots,x_n)^T$ represents a vector of decision variables, and $\Omega \in \mathbb{R}^n$ defines the feasible region within the search space. The objective function $f:\mathbb{R}^n \rightarrow \mathbb{R}$, often considered a black-box due to its intricate internal dynamics, encapsulates many real-world optimization problems that are significantly costly to evaluate. The absence of a closed-form representation of $f(\cdot)$ and the high computational expense associated with its evaluation present considerable challenges for both conventional numerical methods and heuristic optimization algorithms. The primary goal of EAs is to identify the optimal solution $\mathbf{x}^*$ that minimizes the objective function $f(\mathbf{x})$ within a limited number of function evaluations.

\subsection{Surrogate Models assisted Selection}
The high evaluation cost of EOPs presents a significant challenge for EAs. Surrogate models alleviate this challenge by approximating the original function using historical evaluation data through machine learning techniques, thereby reducing the number of calls to the original function during optimization. Various supervised machine learning algorithms are utilized to construct surrogate models, including Gaussian processes \cite{liu2014gaussian}, neural networks \cite{pan2019classificationbased,HaoZZ21}, radial basis function networks \cite{li2020surrogateassisted}, support vector machines \cite{hao2020binary}, and others.

The approximation objectives of surrogate models have evolved from initial function values to include the classification of solutions and the relation of solutions based on their quality, typically categorized into regression models~\cite{jones1998efficient}, classification models~\cite{zhangPreselectionClassificationCase2018,pan2019classificationbased}, and relation models~\cite{hao2024enhancing,hao2022expensive,Hao_2024}. In the context of EAs, the process by which surrogate models actively acquire data for self-updating is known as model management~\cite{jin2018data}. To enhance the accuracy of the surrogate models, numerous model management strategies have been devised. These include generational-based management strategies, where real evaluations are periodically performed over multiple iterations to update the surrogate model~\cite{yaochujin2002framework}, and the selection of representative points using clustering uncertainty and other methods for model updates~\cite{graning2005efficient}. Furthermore, acquisition strategies from Bayesian optimization~\cite{shahriari2015taking}, such as expected improvement (EI)~\cite{mockus1998application} and lower confidence bound (LCB)~\cite{srinivas2009gaussian}, are employed to guide the selection of points for real evaluations and to update the model with new data.

\subsection{Reproduction Operators}

Reproduction operators are fundamental in various EAs for generating new candidate solutions~\cite{wang2023regularity}. In Genetic Algorithm (GA)~\cite{deb1996combined}, selection propagates the fittest individuals, crossover combines parent solutions to produce superior offspring, and mutation introduces variations to maintain diversity. Differential Evolution (DE)~\cite{das2010differential} creates new solutions by combining weighted differences between population vectors, effectively balancing exploration and exploitation. Estimation of Distribution Algorithm (EDA)~\cite{zhouEstimationDistributionAlgorithm2015} sample new candidates based on probabilistic models derived from high-quality solutions, bypassing traditional crossover and mutation. Particle Swarm Optimization (PSO)~\cite{kennedy1995particle} employs social behavior analogies, updating particle positions based on the best-known locations of individuals and their neighbors, facilitating both local and global search. The various reproduction operators in EAs can be uniformly represented by the following equation:
\begin{equation}
    \mathcal{O} = \mathbf{Rep}(\mathcal{P}, N),
    \label{eq:base_reproduction}
\end{equation}
where $\mathcal{O}$ denotes the offspring population, $\mathbf{Rep}(\cdot)$ represents the reproduction operator, $\mathcal{P}$ signifies the parent population, and $N$ is the number of offspring generated.

These operators, specifically designed for EAs, have shown great potential in optimizing performance. However, in the context of SAEAs, their effectiveness is limited by the number of evaluated solutions. This work addresses this limitation by leveraging un-evaluated solutions~$\mathcal{P}_{u}$ to enhance the performance of these reproduction operators.

\section{Motivation}\label{sec:motivation}

\begin{figure}[ht!]
    \centering
    \includegraphics[width=\linewidth]{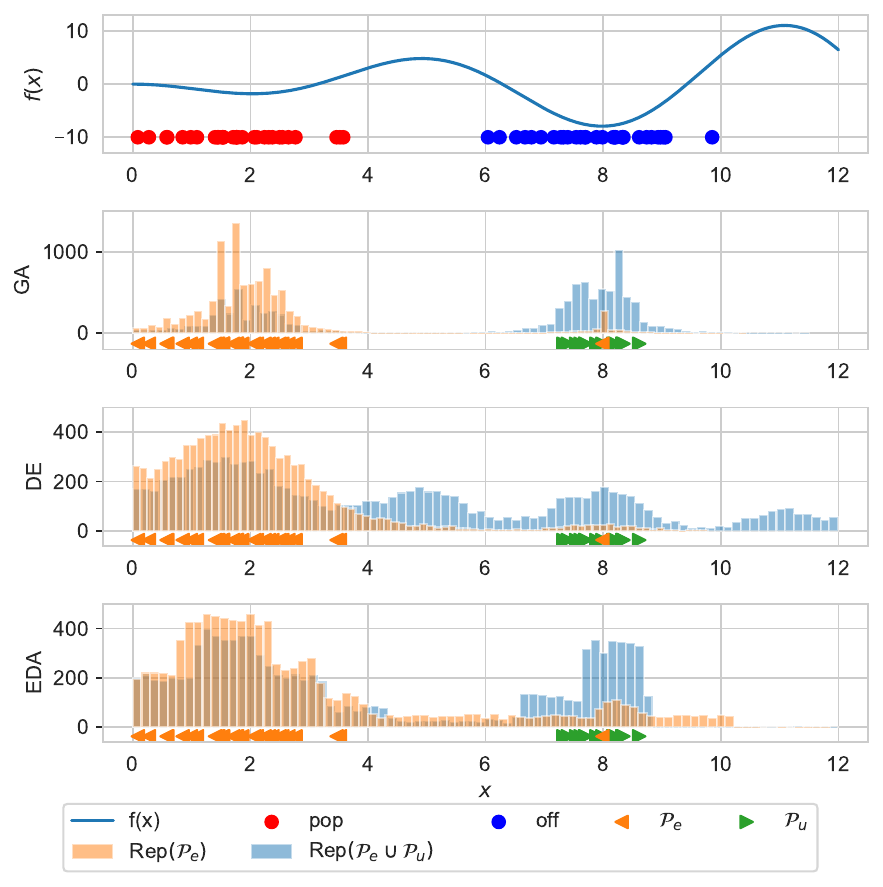}
    \caption{Visualization of the offspring distributions generated by GA, DE, and EDA operators with and without incorporating un-evaluated solutions in one-dimensional function optimization. The orange $\triangleleft$ represent the evaluated population $\mathcal{P}_e$, and the green $\triangleright$ represent the un-evaluated population $\mathcal{P}_u$. The histograms illustrate the offspring distributions generated by the operators.}
    \label{fig:motivation} 
    \end{figure}

In this section, we illustrate the necessity of incorporating un-evaluated solutions through an example involving a 1D function, as shown in Fig~\ref{fig:motivation}. Due to the randomness inherent in operators and EA iterations, a simplified reproduction process is described as follows. Consider a one-dimensional function $f(x) = x\cdot\sin(x)$ defined over the interval $[0, 12]$, with a local optimum in the interval $[0, 4]$ and a global optimum in the interval $[6, 10]$. Assume that at generation $t$, the parent population of size $N = 30$ is located in the interval $[0, 4]$, and the offspring population of size $N$ is situated in the interval $[6, 10]$ (sampled from Gaussian distributions as shown in Fig.~\ref{fig:motivation}). In an extreme scenario, replacing the worst individual in the parent population with the best individual from the offspring population results in a combined population ($\mathcal{P}_e$, denoted by orange $\triangleleft$). Additionally, the top $N/2$ individuals from the offspring population, selected based on their $f(x)$ values, form the un-evaluated population $\mathcal{P}_u$, denoted by green $\triangleright $). In practical algorithm runs, these selections are made according to the surrogate model's predictions to avoid additional evaluation costs. As we proceed to generation $t+1$, the GA, DE, and EDA operators are respectively applied to $\mathcal{P}_e$ and $\mathcal{P}_e \cup \mathcal{P}_u$. Each operator produces 10,000 offspring, and the distributions of these offspring are visualized using histograms. The details of generating offspring using \(\mathcal{P}_e \cup \mathcal{P}_u\) will be discussed in detail in Section~\ref{sec:reproduction}.

While the specific details of the three operators vary, the difference in offspring distribution with and without using un-evaluated solutions is consistent. When using only \(\mathcal{P}_e\), the new generation's distribution does not significantly differ from that of the parent population in generation \(t\). Consequently, few new solutions are found within the optimal region \([6, 10]\). In contrast, when using \(\mathcal{P}_e \cup \mathcal{P}_u\), the offspring distribution contains more solutions in the global optimal region. This demonstrates that incorporating un-evaluated solutions increases population diversity, enhances the operators' performance, and thereby increases the proportion of high-quality solutions generated. Additionally, this approach reduces the difficulty for the surrogate model to select high-quality solutions in subsequent iterations. Therefore, this work advocates incorporating some high-quality solutions into the next generation population without actual evaluations. This approach improves the population's disparity relative to the previous generation, enhances the effectiveness of the operators, and accelerates the optimization process.

\section{Proposed Algorithm}\label{sec:proposed_alg}

In this section, we first present the basic algorithm framework for using un-evaluated solutions assisted evolutionary algorithm (USEA). Next, we provide a detailed explanation of how to integrate un-evaluated solutions with reproduction operators. Finally, we discuss how to train surrogate models and utilize these models to predict solutions.

\subsection{Basic Framework}

\begin{figure}[ht!]
    \centering
    \includegraphics[width=0.9\linewidth]{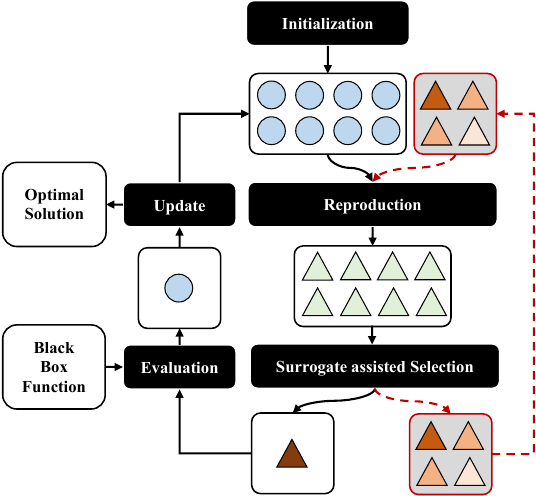}
    \caption{The basic framework of the proposed algorithm. Black arrows connect the fundamental SAEA framework, red arrows indicate our proposed unevaluated solution strategy. Blue circles represent evaluated solutions, green triangles denote offspring, and orange triangles signify solutions assessed by the surrogate model. Darker colors indicate higher quality.}
    \label{fig:framework}
  \end{figure}

The un-evaluated solution strategy proposed in this work is simple and versatile, as shown in Fig.~\ref{fig:framework}. The process connected by black arrows represents a basic surrogate-assisted evolutionary algorithm framework. Starting from initialization, the initial population (circles) is generated. Then, through the reproduction operator, an offspring population (green triangles) is produced. A surrogate model predicts the quality of solutions, and based on these predictions, the most promising solutions~(dark orange triangle) are selected for real evaluation using the black-box function. When the stopping condition of the algorithm is met, it outputs the optimal solution found during iterations.

The process connected by red arrow segments illustrates our proposed un-evaluated solution strategy. In each generation's population, when selecting solutions with surrogate model, some predicted high-quality solutions (light orange triangles; color intensity indicates quality) are chosen and combined with updated next-generation populations to serve as parent population. The reproduction operator then generates an offspring population from these parents. The core challenge of this strategy are how to select high-quality un-evaluated solutions and how to effectively combine them with evaluated solutions to produce offspring solutions.

\begin{algorithm}[ht!]
\caption{Un-evaluated Solutions assisted Evolutionary Algorithm (USEA)}\label{alg:UESA}
\SetKwInOut{Input}{Input}\SetKwInOut{Output}{Output}
\Input{$\mathbf{Rep}$~(reproduction operator),\\$N$~(population size),\\ $\mathcal{M}$ (surrogate model),\\ $\tau$~(training data size),\\ $FEs$~(maximum number of function evaluations).}
\Output{$\mathbf{x}^*$~(optimal solution).}
\BlankLine
$\mathcal{P}_e \leftarrow \mathrm{Initialize}(N)$\; \label{alg:usea:pe-init}
$\mathcal{P}_u \leftarrow \emptyset$ \; \label{alg:usea:pu-init}
$\mathcal{D} \leftarrow \mathcal{P}_e$ \; \label{alg:usea:D-init}
$fes \leftarrow N$\; \label{alg:usea:fes-init}
\While{not $\mathrm{TerminationCondition}(fes,FEs)$}{    \label{alg:usea:while}
    $\mathcal{O} \leftarrow \mathbf{Rep}(\mathcal{P}_e,\mathcal{P}_u,N)$\; \label{alg:usea:gen}
    $\mathcal{M} \leftarrow \mathrm{TrainModel}(\mathcal{D}, \tau)$\;   \label{alg:usea:train}
    $\mathcal{O}^{*},\mathcal{P}_u \leftarrow \mathrm{ModelAssistedSelect}(\mathcal{O}, \mathcal{M})$\;   \label{alg:usea:select}
    $\mathcal{D} \leftarrow \mathcal{D} \cup \mathrm{Evaluate}(\mathcal{O}^{*})$\;    \label{alg:usea:eval}
    $fes \leftarrow fes + |\mathcal{P}_e|$\;    \label{alg:usea:fes-update}
    $\mathcal{P}_e \leftarrow \mathrm{Update}(\mathcal{D},N)$\;     \label{alg:usea:update}
}
\end{algorithm}

Algorithm~\ref{alg:UESA} delineates the basic framework of our proposed USEA. The specific procedural steps are detailed as follows:
\begin{itemize}
  \item \textbf{Initialization} (Lines~\ref{alg:usea:pe-init}-\ref{alg:usea:fes-init}): Initialize a population $\mathcal{P}_e$ of size $N$ using the latin hypercube sampling~(LHS) method~\cite{mckay2000comparisona} and evaluate it using the black-box function $f(\cdot)$. Initialize an empty set $\mathcal{P}_u$ to store un-evaluated solutions. Preserve the evaluated population $\mathcal{P}_e$ in the archive $\mathcal{D}$ and update the evaluation count $fes$ to the population size $N$.
  \item \textbf{Termination Condition} (Line \ref{alg:usea:while}): The termination condition is typically the evaluation count $fes$ reaching the maximum evaluation count $FEs$.
  \item \textbf{New Population Generation} (Line \ref{alg:usea:gen}): Generate a new offspring population $\mathcal{O}$ using the reproduction operator $\mathbf{Rep}$. In the first iteration, $\mathcal{P}_u$ is empty, hence the new population is generated solely from $\mathcal{P}_e$. DE operators, GA operators, and EDA operators can all be utilized as $\mathbf{Rep}(\cdot)$. The integration of these operators with un-evaluated solutions will be discussed in detail in Section~\ref{sec:reproduction}.
  \item \textbf{Model Training} (Line \ref{alg:usea:train}): Train the surrogate model $\mathcal{M}$ using the data from the archive $\mathcal{D}$. The size of the training dataset is denoted by $\tau$. Details of the training method will be elaborated in Section~\ref{sec:surrogate}.
  \item \textbf{Population Selection} (Line \ref{alg:usea:select}): Select the populations $\mathcal{O}^{*}$ and $\mathcal{P}_u$ using the surrogate model $\mathcal{M}$. The specific selection strategies will be thoroughly outlined in Section~\ref{sec:surrogate}.
  \item \textbf{Population Evaluation} (Lines \ref{alg:usea:eval}-\ref{alg:usea:fes-update}): Evaluate the solutions in $\mathcal{O}^{*}$ using the black-box function and add the evaluation results to the archive $\mathcal{D}$. Update the evaluation count $fes$ accordingly.
  \item \textbf{Population Update} (Line \ref{alg:usea:update}): Update the population $\mathcal{P}_e$ using the data in the document $\mathcal{D}$. Based on the real evaluation results, select the best $N$ solutions to form the next generation's population $\mathcal{P}_e$.
\end{itemize}

The following sections provide a detailed explanation on how to effectively integrate un-evaluated solutions with evaluated ones to generate offspring, as well as how to select high-quality un-evaluated solutions. Multiple types of reproduction operators and surrogate models can be incorporated in this unified framework.

\subsection{Reproduction Operator with Un-evaluated Solutions}
\label{sec:reproduction}
This section introduces the integration of un-evaluated solutions into reproduction operators. The reproduction operator for un-evaluated solutions is defined as follows:
\begin{equation}
\mathcal{O} = \mathbf{Rep}(\mathcal{P}_e, \mathcal{P}_u, N),
\label{eq:reproduction}
\end{equation}
where $\mathcal{P}_e$ and $\mathcal{P}_u$ are the evaluated and un-evaluated populations, respectively, and $N$ is the population size. The reproduction operator generates a new population $\mathcal{O}$ based on the evaluated and un-evaluated populations. We will detail the methodologies for combining un-evaluated solutions with GA, DE and EDA operators. For each operator, we first provide the definition of the standard operator and then propose methods for incorporating un-evaluated solutions.

\subsubsection{GA Operators}
In genetic algorithm, the binary tournament selection~\cite{miller1995genetic} is employed to select parent. Subsequently, the simulated binary crossover~(SBX) and polynomial mutation~(PM)~\cite{deb2007selfadaptive} are utilized to generate $N$ offspring solutions. Specifically, the binary tournament selection operates based on the real objective values of the individuals as the pairs of parents. Next, the SBX and PM operators are applied to each pair of parent individuals to perform crossover and mutation operations, yielding two offspring individuals per pair. These offspring individuals are subsequently added to the new population. The final outcome is a new offspring population $\mathcal{O}$ of size $N$. The above three operators are standard. For details, please refer to the supplementary materials.

The method of integrating unevaluated solutions into the GA operators is illustrated in Algorithm~\ref{alg:GA_UE}. The detailed steps are as follows:

\begin{algorithm}[ht!]
\caption{GA operators with un-evaluated solutions}\label{alg:GA_UE}
\SetKwInOut{Input}{Input}\SetKwInOut{Output}{Output}
\Input{$\mathcal{P}_e$~(evaluated population),\\$\mathcal{P}_u$~(un-evaluated population),\\$N$~(population size),\\$\beta_1$~(mixing probability),\\$\beta_2$~(selection probability).}
\Output{$\mathcal{O}$~(offspring population).}
\BlankLine
$\mathcal{O} \leftarrow \emptyset$\; \label{alg:ga:o-init}
\For{$i=1$ \KwTo $N/2$}{
    \tcp{select parents}
    $\{s_e^1, s_e^2\} \leftarrow \mathrm{TournamentSelection}(\mathcal{P}_e, 2)$\;
    $\{s_u^1, s_u^2\} \leftarrow \mathrm{TournamentSelection}(\mathcal{P}_u, 2)$\;
    
    \eIf{$\mathrm{rand}() < \beta_1$}{
        \eIf{$\mathrm{rand}() < \beta_2$}{
            $\{p_1, p_2\} \leftarrow \{s_u^1, s_u^2]$\;
        }{
            \eIf{$\mathrm{rand}() < 0.5$}{
                $\{p_1, p_2\} \leftarrow \{s_e^1, s_u^2\}$\;
            }{
                $\{p_1, p_2\} \leftarrow \{s_u^1, s_e^2\}$\;
            }
        }
    }{
        $\{p_1, p_2\} \leftarrow \{s_e^1, s_e^2\}$\;
    }
    \tcp{crossover and mutation}
    $\{c_1, c_2\} \leftarrow \mathrm{SBX}(p_1, p_2)$\;
    $\{c_1, c_2\} \leftarrow \mathrm{PM}(c_1),\mathrm{PM}(c_2)$\;
    $\mathcal{O} \leftarrow \mathcal{O} \cup \{c_1, c_2\}$\;
}
\end{algorithm}
Algorithm~\ref{alg:GA_UE} incorporates not only the evaluated population $\mathcal{P}_e$ and the un-evaluated population $\mathcal{P}_u$, but also two probability control parameters, $\beta_1$ and $\beta_2$, referred to as the mixing probability and the selection probability, respectively. These parameters govern the preferences for integrating the two populations. Given that the SBX operator generates new solutions based on two parent populations, it is necessary to iterate $N/2$ times to satisfy the requirement of producing $N$ offspring. 

In each iteration, the tournament selection method is employed to select two pairs of solutions from $\mathcal{P}_e$ and $\mathcal{P}_u$ as parent candidates, denoted by $\{s_e^1, s_e^2\}$ and $\{s_u^1, s_u^2\}$, respectively. The selection of $s_e^1$ and $s_e^2$ is based on the evaluated objective value, whereas $s_u^1$ and $s_u^2$ are chosen either based on surrogate model predictions or randomly, as $\mathcal{P}_u$ has been pre-screened by the surrogate model and is considered to contain high-quality solutions. Next, two parent solutions, denoted as $p_1$ and $p_2$, are selected based on the probabilities $\beta_1$ and $\beta_2$. With probability $\beta_1$, a pair of unevaluated solutions are chosen as parents, and with probability $(1-\beta_1)$, a pair of evaluated solutions are used. Subsequently, the method of utilizing the unevaluated solution is determined by $\beta_2$. With probability $(1-\beta_2)$, a solution $s_u$ is randomly selected from the unevaluated parent candidates $\{s_u^1, s_u^2\}$ to replace the evaluated parent solution $s_e^1$ or $s_e^2$; with probability $(\beta_2)$, both unevaluated solutions are used. The SBX operator is then applied to the selected parent solutions to perform crossover, yielding two offspring, $c_1$ and $c_2$. These offspring undergo mutation using the PM operator to generate the final offspring. The offspring are then added to the offspring population $\mathcal{O}$.

Through the aforementioned operations, the GA operator incorporating information from un-evaluated solutions can produce higher-quality offspring.

\subsubsection{DE Operators}
Differential evolution (DE) is a popular evolutionary algorithm that employs mutation and crossover operators. The mutation operator generates a mutant vector by adding the difference between selected individuals to another individual. The crossover operator combines the mutant vector with the target individual ($\mathbf{x}_i$) to produce a trial vector. This work considers five typical DE variants and the specific definitions of these variants are as follows:
\begin{itemize}
    \item rand/1: 
        \begin{equation}
            \mathbf{v}_i = \mathbf{x}_{r_1} + F \cdot (\mathbf{x}_{r_2} - \mathbf{x}_{r_3}) \label{eq:de_rand_1}
        \end{equation}
    \item rand/2: 
        \begin{equation}
            \mathbf{v}_i = \mathbf{x}_{r_1} + F \cdot (\mathbf{x}_{r_2} - \mathbf{x}_{r_3}) + F \cdot (\mathbf{x}_{r_4} - \mathbf{x}_{r_5}) \label{eq:de_rand_2}
        \end{equation}
    \item best/1: 
        \begin{equation}
            \mathbf{v}_i = \mathbf{x}_{\text{best}} + F \cdot (\mathbf{x}_{r_1} - \mathbf{x}_{r_2}) \label{eq:de_best_1}
        \end{equation}
    \item best/2: 
    \begin{equation}
        \mathbf{v}_i = \mathbf{x}_{\text{best}} + F \cdot (\mathbf{x}_{r_1} - \mathbf{x}_{r_2}) + F \cdot (\mathbf{x}_{r_3} - \mathbf{x}_{r_4})  \label{eq:de_best_2}
    \end{equation}
    \item current-to-best/1: 
    \begin{equation}
        \mathbf{v}_i = \mathbf{x}_i + F \cdot (\mathbf{x}_{\text{best}} - \mathbf{x}_i) + F \cdot (\mathbf{x}_{r_1} - \mathbf{x}_{r_2}) \label{eq:de_current_to_best_1}
    \end{equation}
\end{itemize}
In the above equations, $\mathbf{x}_{r_1}, \mathbf{x}_{r_2}, \mathbf{x}_{r_3}, \mathbf{x}_{r_4}, \mathbf{x}_{r_5}$ represent randomly selected individuals, $\mathbf{x}_{\text{best}}$ denotes the best individual in the population, $\mathbf{v}_i$ is the mutant vector, and $F$ is the scaling factor. The ``DE/rand/1'' and ``DE/rand/2'' variants are based on random selection, while the ``DE/best/1'' and ``DE/best/2'' variants utilize the best individual. The ``current-to-best/1'' variant combines the target individual with the best individual. 

Next, the crossover operator is employed to combine the mutant vector with the target individual, producing the trial vector. The specific definition of the crossover operator is defined in equation~(\ref{eq:crossover}):
\begin{equation}
    u_{i,j} = 
    \begin{cases} 
    v_{i,j} & \text{if } \text{rand}_j \leq C_r \text{ or } j = \text{rand}_i \\
    x_{i,j} & \text{otherwise}
    \end{cases}
\label{eq:crossover}
\end{equation}
where $\text{rand}_j$ is a uniformly distributed random number within the range [0, 1], $C_r$ is the crossover rate, and $\text{rand}_i$ is a randomly chosen index to ensure at least one component is taken from the mutant vector. Typically, the operator is applied to the solutions produced by the DE. This additional mutation step aims to further enhance the performance of the operators.

\begin{algorithm}[ht!]
    \caption{DE operators with un-evaluated solutions}\label{alg:DE_UE}
    \SetKwInOut{Input}{Input}\SetKwInOut{Output}{Output}
    \Input{$\mathcal{P}_e$~(evaluated population),\\$\mathcal{P}_u$~(un-evaluated population),\\$N$~(population size),\\$F$~(scaling factor),\\$Cr$~(crossover rate).}
    \Output{$\mathcal{O}$~(offspring population).}
    \BlankLine
    $\mathcal{O} \leftarrow \emptyset$\; \label{alg:de:o-init}
    $\mathcal{P} \leftarrow \mathcal{P}_e \cup \mathcal{P}_u$ \;
    \For{$i=1$ \KwTo $N$}{
        $\mathbf{x}_i, \mathbf{x}_{best} \leftarrow \mathrm{Selection}(\mathcal{P}_e)$\;
        % $\mathbf{x}_{best} \leftarrow \mathrm{Selection}(\mathcal{P}_e)$\;
        $\mathbf{X}_{pool}\leftarrow \mathrm{RandomSelection}(\mathcal{P}/\{\mathbf{x}_{i}\}, 5)$\;
        $\mathbf{v}_{i} \leftarrow \mathrm{Variants}(\mathbf{X}_{pool},\mathbf{x}_{i},\mathbf{x}_{best},F)$ \;
        $\mathbf{u}_{i} \leftarrow \mathrm{Crossover}(\mathbf{x}_{i},\mathbf{v}_{i},Cr)$\;
        $\mathbf{u}_{i} \leftarrow \mathrm{BoundaryCheck}(\mathbf{u}_{i})$\;
        $\mathcal{O} \leftarrow \mathcal{O} \cup \{\mathbf{u}_{i}\}$\;
    }
\end{algorithm}
Algorithm~\ref{alg:DE_UE} demonstrates the incorporation of un-evaluated solutions into DE operators. The principal modification involves selecting the target individual and best individual from $\mathcal{P}_e$. Since solutions in $\mathcal{P}_e$ have been genuinely evaluated, the selection of the best individual is reliable. Conversely, when choosing individuals for the mutant vector, they are randomly selected from the set of $\mathcal{P}_u\cup\mathcal{P}_e$. This strategy ensures that un-evaluated solutions are integrated into the DE operator during the mutation phase. The detailed mutation process follows the definitions provided in equations (3)-(7). Subsequently, the crossover operator is applied to blend the mutant vector with the target individual (equation ), yielding the trial vector. The trial vector then undergoes boundary checking to ensure solution feasibility. Finally, the trial vector is added to the offspring population. Typically, a further PM operation is performed to enhance the performance of the operators~\cite{li2008multiobjective}.

\subsubsection{EDA Operators}

\begin{figure}[ht!]
    \centering
    \includegraphics[width=0.7\linewidth]{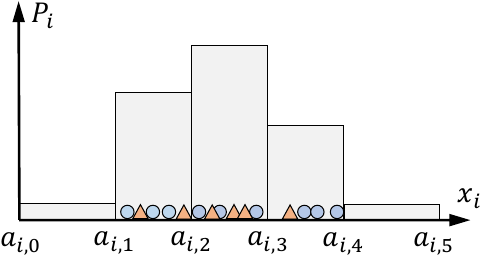}
    \caption{Illustration of the variable width histogram modeling for the $\mathbf{x}_i$ with $K=5$. The first and last bins are assigned very small probabilities due to the absence of individuals falling into them, while bins 2 to 4 are of equal width and model based on the number of individuals within each interval. Here, the blue circles represent the distribution of evaluated solutions ($\mathcal{P}_e$) in the i-th dimension, and the orange triangles indicate the distribution of un-evaluated solutions ($\mathcal{P}_u$) in the i-th dimension.} 
    \label{fig:vwh}
  \end{figure}

EDA operators, incorporated in this work, rely on the variable width histogram (VWH) model proposed in EDA/LS~\cite{zhouEstimationDistributionAlgorithm2015}. The VWH model adheres to the univariate histogram marginal model illustrated by the following equation:
\begin{equation}
    P(\mathbf{x}) =\sum_{i=1}^{n}P_{i}(x_i)
    \label{eq:univariate_model}
\end{equation}
This model constructs a probabilistic model $P_i$ for each dimension $i$ by dividing the search space into $K$ bins, defined as intervals $[a_{i,k}, a_{i,k+1})$, where $k=0, 1,\ldots,K-2$, and the interval $[a_{i,K-1}, a_{i,K}]$. Here, $a_{i,0}$ and $a_{i,K}$ represent the lower bound~($lb_{i}$) and upper bounds~($ub_{i}$), respectively, of the search space for the $i$-th dimension. The smallest and second smallest values in the population for the $i$-th dimension are denoted by $x^1_{i,min}$ and $x^2_{i,min}$, respectively. Similarly, the largest and second largest values are denoted by $x^1_{i,max}$ and $x^2_{i,max}$. To define the widths of the first two bins, we apply the following equations:
\begin{equation}
    a_{i,1} = \max\{x^1_{i,min} - 0.5(x^2_{i,min} - x^1_{i,min}), lb_{i}\}
\end{equation}
\begin{equation}
    a_{i,K-1} = \min\{x^1_{i,max} + 0.5(x^1_{i,max} - x^2_{i,max}), ub_{i}\}
\end{equation}
For bins ranging from the second to the $(K-1)$-th bin, the width is uniform:
\begin{equation}
    a_{i,k} - a_{i,k-1} = \frac{1}{K}(a_{i,K-1} - a_{i,1}), \quad k=2,3,\ldots,K-1
\end{equation}
Upon division into bins along dimension $i$, the probabilistic model $P_i$ can be constructed by calculating the probability of each bin:
\begin{equation}
    P_{i,k} = \frac{N_{i,k}}{\sum_{j=1}^{K}N_{i,j}}
    \label{eq:vwh:prob}
\end{equation}
Here, $N_{i,k}$ refers to the number of individuals in the population whose i-th dimension values fall within the k-th bin. The counting method for $N_k$ is specified as follows:
\begin{equation}
    N_{i,k} = \begin{cases}
    N_{i,k}+1 & \text{if } 1<k<K \\
    0.1 & \text{if } k=1 \text{ or } K \text{ and } a_{i,k} > a_{i,k-1} \\
    0 & \text{if } k=1 \text{ or } K \text{ and } a_{i,k} = a_{i,k-1}
    \end{cases}
\end{equation}
This adjustment ensures that regions without solutions within the search space are still assigned a small probability, thereby maintaining search diversity~\cite{zhouEstimationDistributionAlgorithm2015}. Fig.~\ref{fig:vwh} shows an example of the VWH model for the i-th dimension with $K=5$. Based on the data distribution in the i-th dimension, a histogram model with 5 bins is constructed. In regions where data is distributed, higher probabilities are assigned to accelerate the search process. In regions devoid of data, the histogram model assigns smaller probabilities to ensure diversity in the search.

When generating new solutions, taking the i-th dimension as an example, the first step is to select the k-th bin based on its probability. Subsequently, a new value for the i-th dimension of the solution is uniformly sampled from the interval between $a_{i,k}$ and $a_{i,k+1}$. This sampling process is iteratively repeated until values for all dimensions are obtained. Finally, the newly generated solution is added to the offspring. Furthermore, in EDA/LS, a local search is employed to conduct local searches on the generated solutions, thereby further enhancing the quality of the solutions.

\begin{algorithm}[ht!]
    \caption{EDA operators with un-evaluated solutions}\label{alg:EDA_UE}
    \SetKwInOut{Input}{Input}\SetKwInOut{Output}{Output}
    \Input{$\mathcal{P}_e$~(evaluated population),\\$\mathcal{P}_u$~(un-evaluated population),\\$N$~(population size),\\$K$~(bin size).}
    \Output{$\mathcal{O}$~(offspring population).}
    \BlankLine
    $\mathcal{O} \leftarrow \emptyset$\; \label{alg:eda:o-init}
    $\mathcal{P} \leftarrow \mathcal{P}_e \cup \mathcal{P}_u$ \;
    $P(\mathbf{x}) \leftarrow \mathrm{VWH}(\mathcal{P},K)$ \;
    \For{$i=1$ \KwTo $N$}{
        $\mathbf{u}_{i} \leftarrow \mathrm{Sampling}(P(\mathbf{x}))$\;
        $\mathcal{O} \leftarrow \mathcal{O} \cup \{\mathbf{u}_{i}\}$\;
    }
\end{algorithm}

Algorithm~\ref{alg:EDA_UE} demonstrates the integration of un-evaluated solutions into EDA operators. The algorithm initially merges the evaluated and un-evaluated populations to form a new combined population $\mathcal{P}$. The VWH model is then constructed based on this combined population $\mathcal{P}$ using the specified bin size $K$. Details of the modeling are provided in equations (\ref{eq:univariate_model}) through (\ref{eq:vwh:prob}). Lastly, the probabilistic model $P(X)$ is sampled repetitively $N$ times to generate the offspring population $\mathcal{O}$. 

Algorithm~\ref{alg:GA_UE} through Algorithm~\ref{alg:EDA_UE} demonstrate how un-evaluated solutions are integrated into GA, DE, and EDA operators. As a easy and general strategy, the incorporation of unevaluated solutions is direct. The construction of surrogate models and the selection of un-evaluated solutions will be discussed in detail in Section~\ref{sec:surrogate}.

\subsection{Surrogate Model Training and Selection}
\label{sec:surrogate} 

The training and utilization of surrogate models are also key steps in Algorithm~\ref{alg:UESA}. During the model training phase, an approximate model that reflects the current problem landscape is constructed using historically evaluated data. In the model usage phase, the surrogate model is employed to assess offspring solutions, thereby selecting solutions that require real evaluation, denoted as $\mathcal{O}^*$, as well as un-evaluated solutions $\mathcal{P}_u$. The specific methods for training and using the surrogate model will be elaborated in the following sections.

\subsubsection{Training Surrogate Model}

The data for model training is derived from set $\mathcal{D}$, which contains all the solutions that have been evaluated by the real function during the running of the algorithm, i.e. $\mathcal{D} = \{(\mathbf{x}_i,f(\mathbf{x}_i)),i\in fes\}$. To reduce the size of training data and align the data distribution with current population, we arrange the solutions in ascending order of function value and select the top $\tau$ ones as training data. During the early phase of the USEA algorithm, there might occur situations where $|\mathcal{D}|$ is less than $\tau$. In such cases, all the evaluated solutions are chosen as training data. Thus, a regression model $\mathcal{M}$ that fits the data distribution of the black-box function is constructed, based on the data $\mathcal{D}_{1:\tau}$. as shown in equation~(\ref{eq:train_model}),

\begin{equation}
    y_i = \mathcal{M}(\mathbf{x}_i), \quad i \in \{1,2,\ldots,\tau\}
\label{eq:train_model}
\end{equation}

In this work, we opted for three popular machine learning algorithms as surrogate models, namely, Random Forests~(RF)~\cite{BreimanRandomForests2001}, Gradient Boosting Decision Tree (XGBoost)~\cite{DBLP:conf/kdd/ChenG16}, and Gaussian Process~(GP)~\cite{DBLP:conf/ac/Rasmussen03}. All three models have gained wide adoption in the field of black-box optimization~\cite{hao2020binary,hao2024enhancing} and their performance has been adequately validated. A comprehensive comparison of the performance of these three models will be given in the experimental section.

\subsubsection{Prediction and Selection of Solutions}

\begin{figure}[ht!]
\centering
\includegraphics[width=.5\linewidth]{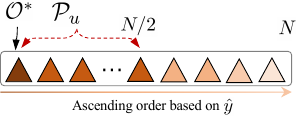}
\caption{Illustration of the surrogate-assisted selection process. The offspring solutions are predicted by the surrogate model, and the top 1 individual is selected as $\mathcal{O}^*$. The top $N/2$ individuals are selected as un-evaluated solutions $\mathcal{P}_u$.}
\label{fig:sas}
\end{figure}

During the model prediction phase, as illustrated in Fig.~\ref{fig:sas}, the offspring $\mathcal{O}$ generated by the reproduction operator will be predicted by the surrogate model $\mathcal{M}$ to obtain the approximate estimated values $\hat{y} = \mathcal{M}(\mathcal{O})$. The predicted values are then sorted in ascending order, and the top 1 individual is selected as $\mathcal{O}^*$, i.e., $\mathcal{O}^* = \arg\min(\hat{y})$. Additionally, based on the predicted values, the top $N/2$ individuals are selected as un-evaluated solutions $\mathcal{P}_u$.

\section{Empirical Studies}\label{sec:experiments}
This section aims to validate the effectiveness of the proposed USEA through a series of experiments. First, we introduce the experimental details, including the experimental setup, comparison algorithms, and test suites. Next, we conduct comparative experiments to evaluate the performance of the USEA algorithm on different test suites. Subsequently, we provide a more detailed analysis, including parameter sensitivity analysis, performance comparison under different surrogate models, ablation studies, and runtime analysis. Finally, a one-dimensional visualization analysis is presented to highlight the differences in the search process between the USEA and Bayesian Optimization (BO) algorithms.

\subsection{Experimental Setup}

\subsubsection{\textbf{Test Suites}}
The test suites include the LZG \cite{liu2014gaussian} and YLL \cite{yao1999evolutionary} test suites. The LZG test suite comprises functions such as Ellipsoid, Rosenbrock, Ackley, and Griewank, which present a variety of landscapes, including unimodal, gully, and multimodal terrains. The YLL test suite, from YLLF01 to YLLF13, features characteristics such as unimodal, multimodal, stepwise, and stochastic noise (excluding YLLF10 and YLLF11 due to overlap with LZG). Considering that the LZG test suite will also be used for subsequent experiments, such as parameter sensitivity testing, the specific details of the functions in the LZG test suite are provided here.

The Ellipsoid function is a convex quadratic function and is unimodal with its global minimum at the origin. The Rosenbrock function is a non-convex function commonly used to evaluate the performance of optimization algorithms, with the global minimum situated within a long, narrow, parabolic-shaped flat valley. The Ackley function is multimodal with a large number of local minima, and its global minimum is located at the origin. The Griewank function is also multimodal, featuring numerous local minima, with the global minimum at the origin. Detailed definitions of the functions in the LZG test suite are provided in the supplementary materials.

\subsubsection{\textbf{Comparison Algorithms}}
For comparative analysis, we identify three representative Bayesian optimization toolkits: the BayesianOptimization toolkit (BOT) \cite{BO2014}, Heteroscedastic Evolutionary Bayesian Optimisation (HEBO) \cite{Cowen-Rivers2022-HEBO}, and Scikit-Optimize (Skopt) \cite{scikit-optimize}. These toolkits are based on Bayesian optimization algorithms and possess the following key attributes:
\begin{itemize}
  \item \textbf{BOT}: Incorporates sequential domain reduction \cite{Standerrobustnesssimpledomain2002a} within the standard BO framework, significantly accelerating the search process and hastening convergence.
  \item \textbf{HEBO}: Implements input and output warping to mitigate the effects of heteroscedasticity and non-stationarity in the objective function. It employs multi-objective optimization to optimize multiple acquisition functions concurrently, thereby enhancing the algorithm's robustness.
  \item \textbf{Skopt}: Utilizes the Hedge strategy \cite{brochu2011portfolio} to probabilistically select among different acquisition functions at each iteration, increasing the robustness of the search process.
\end{itemize}
Additionally, two surrogate-assisted evolutionary algorithms (SAEAs) are utilized in the experiments: the surrogate-assisted multiswarm optimization (SAMSO) \cite{journals/tcyb/LiCGS21} and the fuzzy classification-based preselection for CoDE (FCPS-CoDE) \cite{conf/aaai/ZhouZSZ19}, representing EAs assisted by regression and classification models, respectively. An Estimation of Distribution Algorithm with local search (EDA/LS) \cite{zhouEstimationDistributionAlgorithm2015} serves as the baseline algorithm for this study and is included in the experimental analysis. Detailed descriptions are as follows:
\begin{itemize}
  \item \textbf{SAMSO}: Utilizes a dual-swarm approach with cross-swarm intelligence sharing and dynamic swarm size regulation, guided by a Radial Basis Function (RBF) to estimate the evaluation function.
  \item \textbf{FCPS-CoDE}: Employs a fuzzy K-nearest neighbors (KNN) classifier to predict solution categories before evaluation, aiding the Composite Differential Evolution (CoDE) algorithm in its evolutionary search.
  \item \textbf{EDA/LS}: Integrates an Estimation of Distribution Algorithm (EDA) using a Variable Width Histogram (VWH) and includes a local search operator to facilitate the evolutionary search.
\end{itemize}

\subsubsection{\textbf{Parameter Settings}}
To ensure a fair comparison, the recommended parameters from the original literature are adopted for each algorithm\footnote{The Bayesian optimization packages are used with default settings; SAMSO is implemented within Platemo \cite{tian2017platemo}; FCPS-CoDE and EDA/LS are implemented by us, following the original documentation.}. The specific parameters are as follows:
\begin{itemize}
  \item Maximum umber of evaluations: $FEs = 500$.
  \item Population size: Set to \(N=30\) for EDA/LS and FCPS-CoDE, \(N=40\) for SAMSO, and \(N=50\) for USEA.
  \item Dimensionality: All test problems are set with dimensions \(n=20\) and \(n=50\).
  \item Surrogate model: GP is used for \(n=20\) in the three BO toolkits. For \(n=50\), RFs is preferred due to their lower computational costs. (Note: BOT does not offer an RF option.)
\end{itemize}
The parameter settings for the three operators used in USEA will be detailed in Section~\ref{sec:parameter_sensitivity}. By default, USEA uses RF as the surrogate model.

Each algorithm undergoes 30 independent runs on each test instance to accommodate randomness. The Wilcoxon rank-sum test \cite{hollander2013nonparametric} is applied to analyze the results. In the tables, the symbols `+', `-', and `$\thicksim$' indicate comparative performance against USEA, with `+' denoting significantly better performance, `-' indicating significantly worse, and `$\thicksim$' signifying no significant difference at the 0.05 significance level.

\subsection{Performance Evaluation}
\label{sec:performance_evaluation}

Table~\ref{tab:comparison_results_1} and Table~\ref{tab:comparison_results_2} present the experimental results of USEA combined with three different reproduction operators, where USEA using the EDA operator demonstrates the best performance, followed by the DE operator, and lastly the GA operator. Taking USEA-EDA as the reference and comparing it with BOs and SAEAs, USEA-EDA achieves an mean rank value of 3.53 for $n=20$, performing better than the three EAs and two BOs, but inferior to the HEBO algorithm. The results of the Wilcoxon rank-sum test are largely consistent with the mean rank results, indicating that USEA-EDA is only inferior to the HEBO algorithm. It is worth noting that HEBO uses GP as the surrogate model, incurring higher computational costs (discussed in Section~\ref{sec:runtime_comparison}). For $n=50$ case, where both the HEBO and Skopt algorithms use RF as the surrogate model, USEA-EDA achieves an average rank value of 2.53 and significantly outperforms all other algorithms except BOT in the Wilcoxon rank-sum test. The comparative experimental results indicate that the proposed USEA exhibits strong competitive advantages over mainstream algorithms.

\begin{table*}[ht!]
  \renewcommand{\arraystretch}{1}
  \renewcommand{\tabcolsep}{4pt}
  \centering
  \caption{Statistics of mean and standard deviation results obtained by nine comparison algorithms on LZG and YLL test suites with $n=20$, adhere to a maximum evaluation budget of 500.} 
  \scriptsize
  \begin{tabular}{c|ccc|ccc|ccc}
  \toprule
  \multicolumn{10}{c}{$n=20$} \\
  \midrule
  problem & USEA-EDA & USEA-DE & USEA-GA & HEBO & Skopt & BOT & EDA/LS & SAMSO & FCPS \\
  \midrule
  \multirow{2}{*}{Ellipsoid} & 9.68e+00[4] & 7.32e+01[7]($-$) & 9.49e+01[8]($-$) & 1.41e-01[2]($+$) & \hl{6.58e-02[1]($+$)} & 1.55e-01[3]($+$) & 7.17e+01[6]($-$) & 1.87e+01[5]($\approx$) & 1.30e+02[9]($-$) \\ 
  & (7.14e+00) & (4.46e+01) & (4.26e+01) & (3.84e-02) & (1.75e-02) & (2.83e-02) & (1.56e+01) & (2.52e+01) & (3.13e+01) \\  \hline
  \multirow{2}{*}{Rosenbrock} & 1.04e+02[4] & 1.64e+02[6]($\approx$) & 3.03e+02[8]($-$) & \hl{2.18e+01[1]($+$)} & 5.42e+01[3]($+$) & 1.25e+02[5]($-$) & 2.37e+02[7]($-$) & 3.57e+01[2]($+$) & 3.22e+02[9]($-$) \\ 
  & (2.74e+01) & (1.14e+02) & (1.17e+02) & (1.10e+01) & (1.25e+01) & (4.24e+01) & (4.02e+01) & (2.47e+01) & (1.05e+02) \\  \hline
  \multirow{2}{*}{Ackley} & 7.30e+00[4] & 1.46e+01[6]($-$) & 1.78e+01[8]($-$) & \hl{9.10e-01[1]($+$)} & 7.12e+00[3]($\approx$) & 5.19e+00[2]($+$) & 1.33e+01[5]($-$) & 1.83e+01[9]($-$) & 1.48e+01[7]($-$) \\ 
  & (1.49e+00) & (2.63e+00) & (2.39e+00) & (3.43e-01) & (4.46e-01) & (2.39e+00) & (7.37e-01) & (1.33e+00) & (1.00e+00) \\  \hline
  \multirow{2}{*}{Griewank} & 4.88e+00[4] & 1.88e+01[5]($-$) & 3.12e+01[8]($-$) & \hl{7.82e-01[1]($+$)} & 1.02e+00[2]($+$) & 1.43e+00[3]($+$) & 2.96e+01[7]($-$) & 2.06e+01[6]($-$) & 5.46e+01[9]($-$) \\ 
  & (1.96e+00) & (1.04e+01) & (1.48e+01) & (9.12e-02) & (2.18e-02) & (1.46e-01) & (7.62e+00) & (1.33e+01) & (1.33e+01) \\  \hline
  \multirow{2}{*}{YLLF01} & 3.60e+02[4] & 2.26e+03[6]($-$) & 3.11e+03[7]($-$) & 5.30e+00[2]($+$) & \hl{3.66e+00[1]($+$)} & 1.01e+01[3]($+$) & 3.17e+03[8]($-$) & 6.73e+02[5]($-$) & 5.37e+03[9]($-$) \\ 
  & (1.79e+02) & (1.38e+03) & (1.35e+03) & (1.70e+00) & (1.16e+00) & (2.70e+00) & (5.92e+02) & (6.43e+02) & (1.86e+03) \\  \hline
  \multirow{2}{*}{YLLF02} & \hl{3.32e+00[1]} & 8.06e+00[2]($-$) & 2.22e+01[4]($-$) & 2.04e+01[3]($-$) & 4.62e+05[9]($-$) & 3.89e+04[8]($-$) & 2.67e+01[6]($-$) & 3.27e+01[7]($-$) & 2.41e+01[5]($-$) \\ 
  & (1.03e+00) & (3.21e+00) & (9.66e+00) & (6.62e+00) & (9.03e+05) & (1.21e+05) & (3.93e+00) & (1.72e+01) & (3.67e+00) \\  \hline
  \multirow{2}{*}{YLLF03} & 8.49e+03[3] & \hl{6.37e+03[1]($+$)} & 1.67e+04[5]($-$) & 1.68e+04[6]($-$) & 7.47e+03[2]($\approx$) & 3.50e+04[9]($-$) & 2.32e+04[8]($-$) & 1.71e+04[7]($\approx$) & 1.26e+04[4]($-$) \\ 
  & (1.96e+03) & (3.59e+03) & (5.16e+03) & (4.33e+03) & (2.69e+03) & (6.98e+03) & (4.21e+03) & (1.38e+04) & (3.48e+03) \\  \hline
  \multirow{2}{*}{YLLF04} & 3.85e+01[5] & 3.88e+01[6]($\approx$) & 6.44e+01[8]($-$) & \hl{2.14e+01[1]($+$)} & 2.78e+01[2]($+$) & 2.82e+01[3]($+$) & 3.28e+01[4]($+$) & 7.06e+01[9]($-$) & 4.17e+01[7]($\approx$) \\ 
  & (8.46e+00) & (1.01e+01) & (8.68e+00) & (1.14e+01) & (1.72e+01) & (9.43e+00) & (3.26e+00) & (7.40e+00) & (6.55e+00) \\  \hline
  \multirow{2}{*}{YLLF05} & 7.95e+04[2] & 3.74e+05[5]($-$) & 4.43e+06[8]($-$) & \hl{3.90e+02[1]($+$)} & 1.77e+05[4]($-$) & 2.34e+06[7]($-$) & 1.13e+06[6]($-$) & 1.12e+05[3]($\approx$) & 4.71e+06[9]($-$) \\ 
  & (6.58e+04) & (3.91e+05) & (4.06e+06) & (6.20e+02) & (8.96e+04) & (1.04e+06) & (5.76e+05) & (8.27e+04) & (2.84e+06) \\  \hline
  \multirow{2}{*}{YLLF06} & 4.78e+02[4] & 2.06e+03[6]($-$) & 4.12e+03[8]($-$) & 6.97e+00[2]($+$) & \hl{5.30e+00[1]($+$)} & 9.80e+00[3]($+$) & 3.20e+03[7]($-$) & 8.59e+02[5]($-$) & 6.26e+03[9]($-$) \\ 
  & (4.67e+02) & (1.32e+03) & (2.54e+03) & (2.09e+00) & (1.53e+00) & (3.19e+00) & (5.04e+02) & (1.24e+03) & (1.28e+03) \\  \hline
  \multirow{2}{*}{YLLF07} & 3.07e-01[3] & 1.99e+00[8]($-$) & 1.33e+00[7]($-$) & \hl{9.19e-02[1]($+$)} & 2.67e-01[2]($\approx$) & 1.05e+00[6]($-$) & 6.67e-01[5]($-$) & 3.23e-01[4]($\approx$) & 2.09e+00[9]($-$) \\ 
  & (1.52e-01) & (1.35e+00) & (9.03e-01) & (3.87e-02) & (9.91e-02) & (4.24e-01) & (2.75e-01) & (1.56e-01) & (9.17e-01) \\  \hline
  \multirow{2}{*}{YLLF08} & 2.21e+03[2] & 2.39e+03[3]($\approx$) & 3.02e+03[4]($-$) & \hl{1.86e+03[1]($+$)} & 5.17e+03[6]($-$) & 5.54e+03[8]($-$) & 4.66e+03[5]($-$) & 5.67e+03[9]($-$) & 5.52e+03[7]($-$) \\ 
  & (6.02e+02) & (5.87e+02) & (4.57e+02) & (3.48e+02) & (3.29e+02) & (5.52e+02) & (3.32e+02) & (2.83e+02) & (2.88e+02) \\  \hline
  \multirow{2}{*}{YLLF09} & 1.26e+02[5] & 1.03e+02[3]($+$) & 1.29e+02[6]($\approx$) & \hl{8.66e+01[1]($+$)} & 1.63e+02[8]($-$) & 1.65e+02[9]($-$) & 1.57e+02[7]($-$) & 1.16e+02[4]($\approx$) & 1.02e+02[2]($+$) \\ 
  & (1.77e+01) & (4.04e+01) & (2.21e+01) & (1.96e+01) & (1.86e+01) & (3.53e+01) & (1.31e+01) & (5.14e+01) & (2.09e+01) \\  \hline
  \multirow{2}{*}{YLLF12} & 3.28e+03[3] & 1.39e+05[6]($-$) & 4.06e+06[8]($-$) & \hl{2.47e+00[1]($+$)} & 6.68e+04[5]($-$) & 8.77e+06[9]($-$) & 6.09e+04[4]($-$) & 1.23e+02[2]($+$) & 2.46e+06[7]($-$) \\ 
  & (6.75e+03) & (2.98e+05) & (8.60e+06) & (1.61e+00) & (1.21e+05) & (5.76e+06) & (8.76e+04) & (5.49e+02) & (3.04e+06) \\  \hline
  \multirow{2}{*}{YLLF13} & 1.95e+10[5] & 1.58e+11[7]($-$) & 5.01e+11[8]($-$) & \hl{3.17e+04[1]($+$)} & 2.72e+10[6]($-$) & 8.77e+11[9]($-$) & 1.10e+06[3]($+$) & 6.82e+04[2]($+$) & 1.08e+07[4]($+$) \\ 
  & (2.07e+10) & (2.08e+11) & (5.30e+11) & (6.92e+04) & (1.23e+10) & (2.20e+11) & (5.84e+05) & (1.94e+05) & (8.07e+06) \\  \hline
  mean rank & 3.53 & 5.13 & 7.00 & 1.67 & 3.67 & 5.8 & 5.87 & 5.27 & 7.07 \\ 
  $+$ / $-$ / $\approx$ & & 2/10/3 & 0/14/1 & 13/2/0 & 6/6/3 & 6/9/0 & 2/13/0 & 3/7/5 & 2/12/1 \\ 
  \bottomrule
  \end{tabular}
  \label{tab:comparison_results_1}
\end{table*}

\begin{table*}[ht!]
  \renewcommand{\arraystretch}{1.1}
  \renewcommand{\tabcolsep}{4pt}
\centering
\caption{Statistics of mean and standard deviation results obtained by nine comparison algorithms on LZG and YLL test suites with $n=50$, adhere to a maximum evaluation budget of 500.} 
\scriptsize
\begin{tabular}{c|ccc|ccc|ccc}
\toprule
\multicolumn{10}{c}{$n=50$} \\
\midrule
problem & USEA-EDA & USEA-DE & USEA-GA & HEBO & Skopt & BOT & EDA/LS & SAMSO & FCPS \\
\midrule
\multirow{2}{*}{Ellipsoid} & 6.80e+02[2] & 1.85e+03[6]($-$) & 3.13e+03[7]($-$) & 3.59e+03[8]($-$) & 5.87e+03[9]($-$) & \hl{7.52e+00[1]($+$)} & 1.52e+03[4]($-$) & 1.31e+03[3]($\approx$) & 1.61e+03[5]($-$) \\ 
& (1.28e+02) & (6.07e+02) & (5.50e+02) & (4.78e+02) & (4.37e+02) & (3.19e+00) & (2.23e+02) & (1.13e+03) & (3.39e+02) \\  \hline
\multirow{2}{*}{Rosenbrock} & 9.00e+02[2] & 2.68e+03[6]($-$) & 3.97e+03[7]($-$) & 5.26e+03[8]($-$) & 9.59e+03[9]($-$) & \hl{5.02e+02[1]($+$)} & 1.78e+03[4]($-$) & 1.58e+03[3]($\approx$) & 1.90e+03[5]($-$) \\ 
& (1.68e+02) & (7.96e+02) & (9.05e+02) & (7.33e+02) & (1.03e+03) & (1.10e+02) & (2.84e+02) & (1.70e+03) & (5.11e+02) \\  \hline
\multirow{2}{*}{Ackley} & 1.53e+01[2] & 1.89e+01[6]($-$) & 1.99e+01[7]($-$) & 2.00e+01[8]($-$) & 2.06e+01[9]($-$) & \hl{8.36e+00[1]($+$)} & 1.76e+01[4]($-$) & 1.86e+01[5]($-$) & 1.75e+01[3]($-$) \\ 
& (8.31e-01) & (6.82e-01) & (2.24e-01) & (2.71e-01) & (1.40e-01) & (6.22e-01) & (4.24e-01) & (1.18e+00) & (6.00e-01) \\  \hline
\multirow{2}{*}{Griewank} & 1.17e+02[2] & 2.96e+02[5]($-$) & 4.75e+02[6]($-$) & 5.52e+02[7]($-$) & 8.83e+02[9]($-$) & \hl{3.74e+00[1]($+$)} & 2.41e+02[3]($-$) & 6.19e+02[8]($-$) & 2.69e+02[4]($-$) \\ 
& (2.03e+01) & (7.11e+01) & (7.43e+01) & (7.86e+01) & (5.97e+01) & (5.77e-01) & (2.86e+01) & (3.66e+02) & (6.72e+01) \\  \hline
\multirow{2}{*}{YLLF01} & 1.31e+04[2] & 3.38e+04[5]($-$) & 5.08e+04[6]($-$) & 6.17e+04[7]($-$) & 9.56e+04[9]($-$) & \hl{6.58e+01[1]($+$)} & 2.77e+04[3]($-$) & 6.69e+04[8]($-$) & 2.95e+04[4]($-$) \\ 
& (2.71e+03) & (9.35e+03) & (6.90e+03) & (9.04e+03) & (8.96e+03) & (8.57e+00) & (3.92e+03) & (3.70e+04) & (5.86e+03) \\  \hline
\multirow{2}{*}{YLLF02} & \hl{4.23e+01[1]} & 4.83e+01[2]($-$) & 2.26e+08[6]($-$) & 3.31e+04[5]($-$) & 8.70e+17[8]($-$) & 3.06e+17[7]($-$) & 2.79e+04[4]($-$) & 1.05e+18[9]($-$) & 8.92e+01[3]($-$) \\ 
& (3.47e+00) & (1.18e+01) & (7.21e+08) & (1.71e+05) & (4.40e+18) & (1.18e+18) & (9.04e+04) & (3.77e+18) & (8.24e+00) \\  \hline
\multirow{2}{*}{YLLF03} & 9.59e+04[3] & \hl{5.20e+04[1]($+$)} & 1.07e+05[4]($\approx$) & 1.33e+05[5]($-$) & 1.83e+05[7]($-$) & 1.85e+05[8]($-$) & 1.58e+05[6]($-$) & 2.97e+05[9]($-$) & 8.24e+04[2]($+$) \\ 
& (1.90e+04) & (1.74e+04) & (2.21e+04) & (2.57e+04) & (3.44e+04) & (3.21e+04) & (2.51e+04) & (9.31e+04) & (1.68e+04) \\  \hline
\multirow{2}{*}{YLLF04} & 6.28e+01[4] & 6.17e+01[3]($\approx$) & 8.30e+01[7]($-$) & 8.08e+01[6]($-$) & 9.05e+01[9]($-$) & 7.28e+01[5]($-$) & \hl{5.80e+01[1]($+$)} & 8.74e+01[8]($-$) & 5.88e+01[2]($+$) \\ 
& (3.82e+00) & (8.69e+00) & (4.11e+00) & (4.20e+00) & (2.23e+00) & (1.52e+01) & (2.78e+00) & (8.06e+00) & (5.58e+00) \\  \hline
\multirow{2}{*}{YLLF05} & 1.03e+07[2] & 4.09e+07[4]($-$) & 1.32e+08[6]($-$) & 1.74e+08[8]($-$) & 3.36e+08[9]($-$) & \hl{8.64e+06[1]($+$)} & 2.93e+07[3]($-$) & 1.55e+08[7]($-$) & 4.49e+07[5]($-$) \\ 
& (2.65e+06) & (1.88e+07) & (3.30e+07) & (4.91e+07) & (3.76e+07) & (3.24e+06) & (5.30e+06) & (8.04e+07) & (1.82e+07) \\  \hline
\multirow{2}{*}{YLLF06} & 1.20e+04[2] & 3.51e+04[5]($-$) & 4.99e+04[6]($-$) & 6.22e+04[7]($-$) & 9.65e+04[9]($-$) & \hl{7.19e+01[1]($+$)} & 2.74e+04[3]($-$) & 7.82e+04[8]($-$) & 2.92e+04[4]($-$) \\ 
& (2.45e+03) & (1.02e+04) & (8.59e+03) & (7.44e+03) & (7.35e+03) & (1.43e+01) & (4.41e+03) & (3.06e+04) & (7.35e+03) \\  \hline
\multirow{2}{*}{YLLF07} & \hl{7.86e+00[1]} & 3.63e+01[5]($-$) & 8.89e+01[6]($-$) & 1.25e+02[7]($-$) & 2.56e+02[9]($-$) & 8.35e+00[2]($\approx$) & 2.17e+01[3]($-$) & 1.31e+02[8]($-$) & 2.96e+01[4]($-$) \\ 
& (3.11e+00) & (1.62e+01) & (2.25e+01) & (3.54e+01) & (3.52e+01) & (2.96e+00) & (4.75e+00) & (7.91e+01) & (1.28e+01) \\  \hline
\multirow{2}{*}{YLLF08} & 1.39e+04[4] & 1.15e+04[3]($+$) & 1.09e+04[2]($+$) & \hl{1.09e+04[1]($+$)} & 1.58e+04[6]($-$) & 1.69e+04[9]($-$) & 1.51e+04[5]($-$) & 1.66e+04[8]($-$) & 1.63e+04[7]($-$) \\ 
& (8.26e+02) & (2.42e+03) & (7.92e+02) & (8.40e+02) & (6.43e+02) & (5.53e+02) & (5.84e+02) & (6.62e+02) & (6.11e+02) \\  \hline
\multirow{2}{*}{YLLF09} & 5.17e+02[5] & 3.96e+02[2]($+$) & 4.53e+02[3]($+$) & 5.28e+02[8]($\approx$) & 6.98e+02[9]($-$) & 5.18e+02[6]($\approx$) & 5.19e+02[7]($\approx$) & 4.97e+02[4]($\approx$) & \hl{3.80e+02[1]($+$)} \\ 
& (2.98e+01) & (9.27e+01) & (3.90e+01) & (3.68e+01) & (3.04e+01) & (3.64e+01) & (1.78e+01) & (1.11e+02) & (3.15e+01) \\  \hline
\multirow{2}{*}{YLLF12} & \hl{4.46e+06[1]} & 5.51e+07[5]($-$) & 1.99e+08[6]($-$) & 3.20e+08[7]($-$) & 7.46e+08[9]($-$) & 2.10e+07[3]($-$) & 1.64e+07[2]($-$) & 5.10e+08[8]($-$) & 5.44e+07[4]($-$) \\ 
& (2.40e+06) & (3.35e+07) & (7.11e+07) & (1.17e+08) & (1.24e+08) & (1.73e+07) & (5.98e+06) & (3.16e+08) & (3.49e+07) \\  \hline
\multirow{2}{*}{YLLF13} & 1.68e+12[5] & 6.35e+12[6]($-$) & 1.98e+13[7]($-$) & 2.81e+13[8]($-$) & 5.42e+13[9]($-$) & 1.39e+12[4]($\approx$) & \hl{7.53e+07[1]($+$)} & 8.32e+08[3]($+$) & 1.58e+08[2]($+$) \\ 
& (6.69e+11) & (3.41e+12) & (4.86e+12) & (7.18e+12) & (6.88e+12) & (6.90e+11) & (2.19e+07) & (6.39e+08) & (1.01e+08) \\  \hline
mean rank & 2.53 & 4.26 & 5.67 & 6.67 & 8.60 & 3.40 & 3.53 & 6.60 & 3.67 \\ 
$+$ / $-$ / $\approx$ &  & 3/11/1 & 2/12/1 & 1/13/1 & 0/15/0 & 7/5/3 & 2/12/1 & 1/11/3 & 4/11/0 \\ 
\bottomrule
\end{tabular}
\label{tab:comparison_results_2}
\end{table*}

\subsection{Parameter Sensitivity Analysis}
\label{sec:parameter_sensitivity}
Although incorporating un-evaluated solutions with traditional reproduction operators is straightforward, there are still several parameters and strategies in Algorithm~\ref{alg:GA_UE}–\ref{alg:EDA_UE} that require fine-tuning. These include the $\beta_1$ and $\beta_2$ parameters in the GA operator, which control the combination probabilities, the choice of various variants in the DE operator, and the bin size $K$ in the EDA operator. This section will conduct a sensitivity analysis of these parameters.

We selected the LZG test suite and set the problem dimensions to 20 and 50. We analyzed the parameter sensitivity for the GA, DE, and EDA operators. Due to the significant variance in final solution results value across different test problems, we use an improvement metric as an evaluation indicator, defined in equation~(\ref{eq:improvement})
\begin{equation}
    I = \frac{\mathrm{baseline}-\mathrm{variant}}{\mathrm{baseline}} \times 100\%
    \label{eq:improvement}
\end{equation}
where `$\mathrm{baseline}$' represents the mean result obtained by the original operators without using un-evaluated solutions within the USEA framework, solved independently 30 times for each test suite. In contrast, `$\mathrm{variant}$' refers to the mean results obtained by operators with different parameter settings using un-evaluated solutions. A larger improvement value $I$ indicates better operator performance. 

Fig.~\ref{fig:usea_ga}–\ref{fig:usea_eda} illustrate radar charts showing the improvement metrics of the GA, DE and EDA operators on the LZG test suite, respectively. The outer ring scale, labeled 1–4, represents the improvement metrics for the Ellipsoid, Rosenbrock, Ackley, and Griewank test problems in 20 dimensions, while 5–8 represent the metrics for these four test problems in 50 dimensions. The performance of the baseline ($I=0$) is indicated by a black dashed line. More detailed experimental data can be found in the supplementary materials.

\subsubsection{Study of $\beta_1$ and $\beta_2$ in GA operator}

\begin{figure}[ht!]
\centering
\includegraphics[width=1\linewidth]{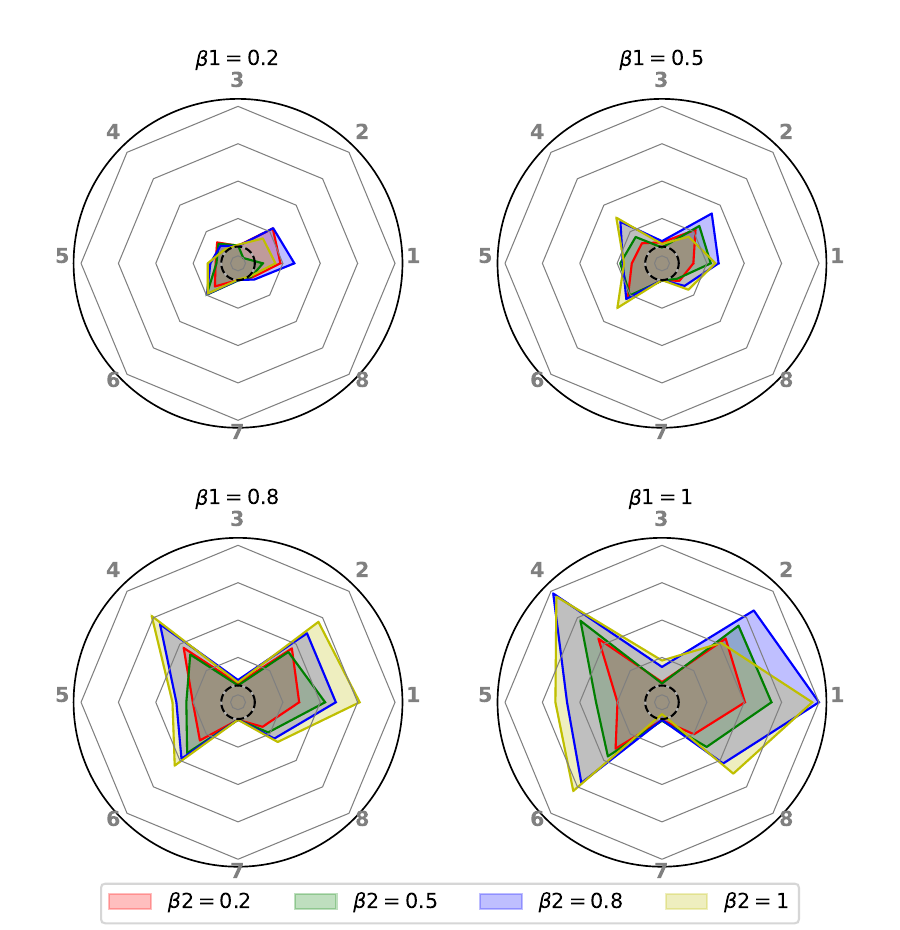}
\caption{Performance improvement of the  GA operator with different $\beta_1$ and $\beta_2$ values of un-evaluated solutions on the LZG test suite.}
\label{fig:usea_ga}
\end{figure}

The parameters $\beta_1$ and $\beta_2$ respectively control the probability of the un-evaluated solution participating in the generation of offspring in the GA operator. Each parameter takes values of 0.2, 0.5, 0.8, and 1. Fig.~\ref{fig:usea_ga} illustrates the performance of the GA operator with un-evaluated solutions under different parameter combinations. From the figure, it can be observed that as $\beta_1$ increases, the performance of the GA operator gradually improves, indicating that increasing the number of un-evaluated solutions enhances the performance of the GA operator. When $\beta_2$ is greater than 0.5, the GA operator performs comparatively well, suggesting that using evaluated and un-evaluated solutions respectively as parents to generate offspring is more effective than directly using un-evaluated solutions for new solution generation. Additionally, from the position of the dashed line, it is evident that when $\beta_1$ is greater than 0.5, the performance of all variants surpasses the baseline, indicating the robustness of performance improvement due to the inclusion of un-evaluated solutions. Moreover, regarding the 8 scales, the Ackley function shows weak performance improvement both in 20 dimensions and 50 dimensions (scales 3 and 7). Finally, we select $\beta_1 = 1$ and $\beta_2 = 0.8$ as the optimal parameter combination for the GA operator.

\subsubsection{Study of variants in DE operator}

\begin{figure}[ht!]
\centering
\includegraphics[width=0.7\linewidth]{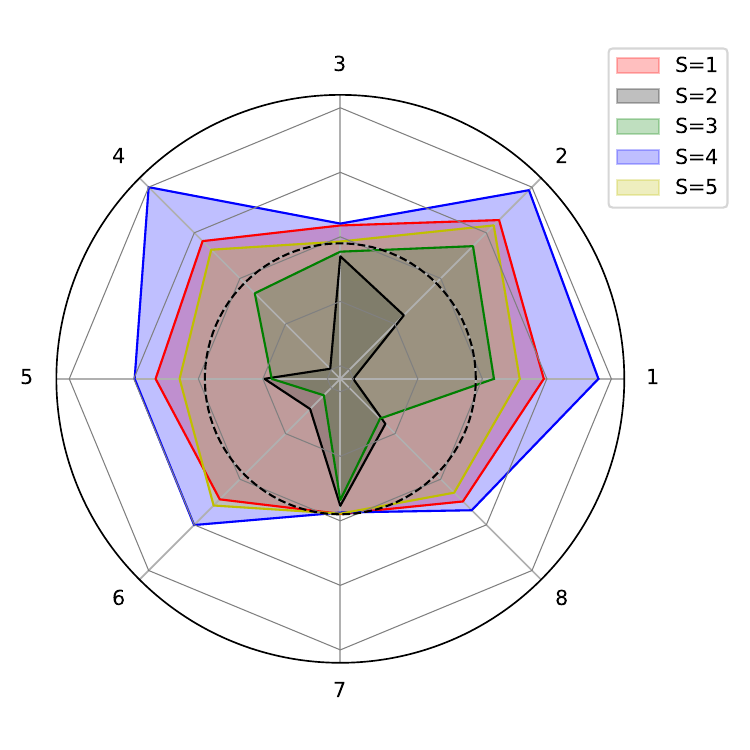}
\caption{Performance improvement of the DE operator with different variants with un-evaluated solutions on the LZG test suite. The legend indicates the DE variants referred to in equations (\ref{eq:de_rand_1})–(\ref{eq:de_current_to_best_1}).}
\label{fig:usea_de}
\end{figure}

This section analyzes the performance of different DE variants combined with un-evaluated solutions. Specifically, the DE variants include ``rand/1'', ``rand/2'', ``best/1'', ``best/2'', and ``current-to-best/1'', as described in equations (\ref{eq:de_rand_1})–(\ref{eq:de_current_to_best_1}), denoted as $S=1,\ldots,5$, respectively. The parameters $F$ and $Cr$ for each DE variant are set to 0.5 and 0.9~\cite{tian2017platemo}, respectively. The baseline represents the performance of the DE operator without using un-evaluated solutions and using the ``rand/1'' strategy. Fig.~\ref{fig:usea_de} shows the performance improvement of these DE variants. From the figure, it can be observed that the ``best/2'' strategy ($S=4$) shows the best performance, followed by the ``rand/1'' strategy ($S=1$). Notably, both $S=1$ and the baseline use the same DE variant, but the performance of $S=1$ significantly surpasses the baseline, indicating that the introduction of un-evaluated solutions has a significant positive impact on the DE operator's performance. The $S=2$ and $S=3$ strategies perform relatively poorly. Considering these results comprehensively, we select the `best/2' strategy as the optimal strategy for the DE operator combined with un-evaluated solutions.

\subsubsection{Study of bin size $K$ in EDA operator}

\begin{figure}[ht!]
    \centering
    \includegraphics[width=0.7\linewidth]{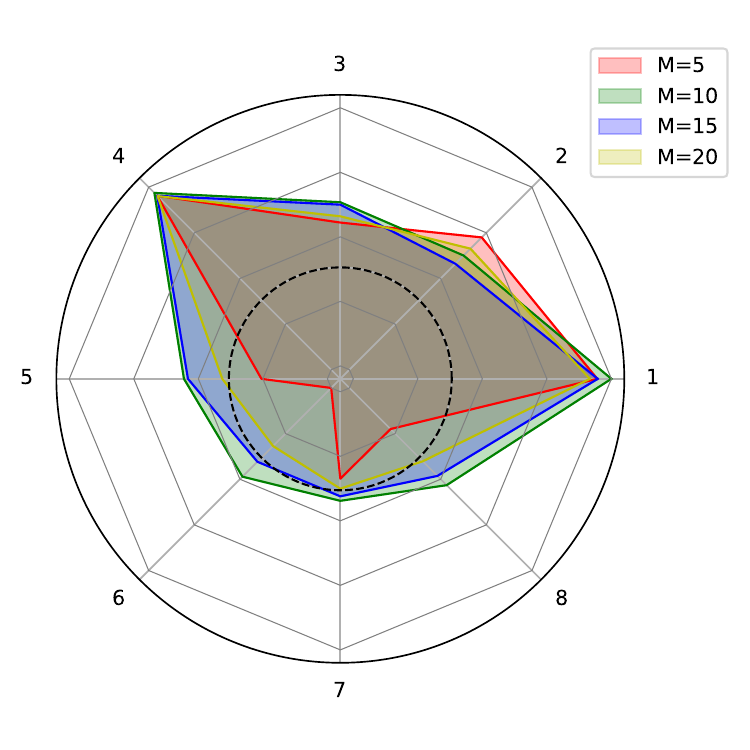}
    \caption{Performance improvement of the EDA operator with different bin sizes $K$ with un-evaluated solutions on the LZG test suite.}
    \label{fig:usea_eda}
  \end{figure}

The number of bins $K$ in the EDA operator is a critical parameter that affects the modeling accuracy of VWH. We set $K$ to 5, 10, 15, and 20, respectively, and analyze the performance of the EDA operator with un-evaluated solutions on the LZG test suite. The baseline is set as the EDA operator without using un-evaluated solutions, with $K$ set to 15 ~\cite{zhouEstimationDistributionAlgorithm2015}. Fig.~\ref{fig:usea_eda} shows the performance of the EDA operator with different bin sizes $K$. From the figure, it can be observed that for problem dimensions of 20 (outer ring scale 1–4), the EDA operator with un-evaluated solutions performs better than the baseline, regardless of the value of $K$. For problem dimensions of 50 (outer ring scale 5–8), performance is poor for $K=5$ and $K=20$, and in some cases, it is worse than the baseline. However, for $K=10$ and $K=15$, performance is better and exceeds the baseline. Considering these results comprehensively, we select $K=10$ as the optimal parameter setting for the EDA operator.

\subsection{Comparison of Surrogate Models}
This section discusses the impact of different surrogate models on the performance of the USEA algorithm. We selected RF, GP, and XGB as surrogate models and conducted experiments on the LZG test suite. Table~\ref{tab:surrogate_results} presents the statistical results after 30 independent runs of the USEA using different surrogate models.

The experimental results indicate that for \(n=20\), both GP and XGB models can further improve the performance of the USEA algorithm compared to RF. For \(n=50\), GP significantly outperforms RF on two test problems but performs significantly worse on the other two, while XGB performs slightly worse than RF overall. Based on these considerations, RF was selected as the default surrogate model for USEA.

\begin{table}[H]
\renewcommand{\arraystretch}{1.1}
\renewcommand{\tabcolsep}{10pt}
\centering
\caption{Statistics of mean and standard deviation results obtained by USEA with RF, GP and XGB on LZG test suite.} \scriptsize
\begin{tabular}{cccc}
\toprule
\multicolumn{4}{c}{$n=20$} \\
\midrule
problem & USEA-RF & USEA-GP & USEA-XGB \\
\midrule
\multirow{2}{*}{Ellipsoid} & 9.68e+00[3] & \hl{1.42e-02[1]($+$)} & 5.80e+00[2]($\approx$) \\ 
    & (7.14e+00) & (1.31e-02) & (2.22e+00) \\  \hline
\multirow{2}{*}{Rosenbrock} & 1.04e+02[3] & \hl{5.79e+01[1]($+$)} & 8.30e+01[2]($+$) \\ 
    & (2.74e+01) & (2.73e+01) & (3.29e+01) \\  \hline
\multirow{2}{*}{Ackley} & 7.30e+00[3] & \hl{4.87e+00[1]($+$)} & 7.27e+00[2]($\approx$) \\ 
    & (1.49e+00) & (1.82e+00) & (9.97e-01) \\  \hline
\multirow{2}{*}{Griewank} & 4.88e+00[2] & 1.78e+01[3]($-$) & \hl{3.06e+00[1]($+$)} \\ 
    & (1.96e+00) & (3.99e+00) & (9.26e-01) \\  \hline
mean rank & 2.75 & 1.50 & 1.75 \\ 
$+$ / $-$ / $\approx$ & & 3/1/0 & 2/0/2 \\ 
\midrule
\multicolumn{4}{c}{$n=50$} \\
\midrule
problem & USEA-RF & USEA-GP & USEA-XGB \\
\midrule
\multirow{2}{*}{Ellipsoid} & 6.80e+02[2] & \hl{2.22e+02[1]($+$)} & 7.96e+02[3]($-$) \\ 
& (1.28e+02) & (5.30e+01) & (1.45e+02) \\  \hline
\multirow{2}{*}{Rosenbrock} & 9.00e+02[3] & \hl{6.37e+02[1]($+$)} & 8.91e+02[2]($\approx$) \\ 
& (1.68e+02) & (1.24e+02) & (1.81e+02) \\  \hline
\multirow{2}{*}{Ackley} & \hl{1.53e+01[1]} & 1.58e+01[3]($-$) & 1.57e+01[2]($-$) \\ 
& (8.31e-01) & (5.15e-01) & (5.94e-01) \\  \hline
\multirow{2}{*}{Griewank} & \hl{1.17e+02[1]} & 1.44e+02[3]($-$) & 1.26e+02[2]($\approx$) \\ 
& (2.03e+01) & (2.03e+01) & (2.29e+01) \\  \hline
mean rank & 1.75 & 2.00 & 2.25 \\ 
$+$ / $-$ / $\approx$ && 2/2/0 & 0/2/2 \\ 
\bottomrule
\end{tabular}
\label{tab:surrogate_results}
\end{table}

\subsection{Ablation Study}
\label{sec:ablation_study}
This section substantiates the effectiveness of the presented USEA through ablation studies. Variants are described follow: 
\begin{itemize}
    \item USEA: USEA with RF surrogate.
    \item USEA-AL: Evaluate all solutions selected by the model.
    \item USEA-NS: Excludes unevaluated solutions from reproduction.
    \item EDA/LS~\cite{zhouEstimationDistributionAlgorithm2015}: The baseline algorithm for USEA.
\end{itemize}

\begin{figure*}
    \centering
    \includegraphics[width=0.99\linewidth]{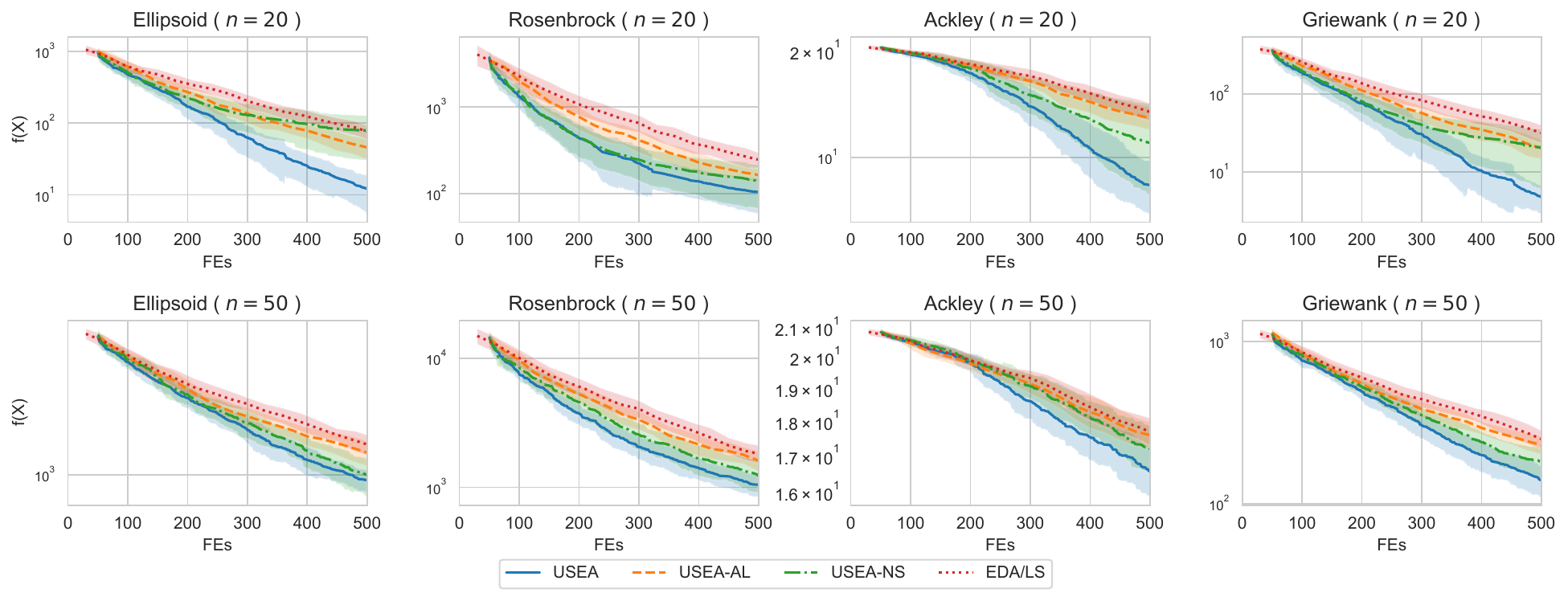}
    \caption{Ablation study of USEA on the LZG test suite with $n=20,50$.}
    \label{fig:ablation}
\end{figure*}

The ablation experiments examine the benefits of USEA compared to the baseline EDA/LS, as well as the impact of evaluating all surrogate model-selected solutions and excluding unevaluated solutions in reproduction. Conducted on the LZG test suite with problem dimensions of \(n=20, 50\), each experimental set was independently replicated 30 times to account for randomness. Fig.~\ref{fig:ablation} illustrates the objective values versus the number of fitness evaluations (FEs), with lines representing mean values and shaded areas indicating standard deviation.

The results clearly indicate that USEA consistently outperforms both its variants and the baseline across all problems, demonstrating the efficacy of USEA and the importance of surrogate-assisted selection and reproduction strategies. Among the variants, USEA-AL performs better than USEA-NS, and both surpass the baseline algorithm. Detailed analysis reveals that EDA/LS's fixed evaluation of a set number of individuals per generation is unsuitable for expensive optimization problems, as it leads to high computational expense. With limited FEs (500), the search process is insufficient. USEA-ALL, by evaluating only half the individuals selected by the surrogate model per generation, effectively increases the number of evolutionary generations, thereby achieving better performance. Conversely, USEA-NS, which evaluates one individual per generation to save computational resources, loses valuable information from unevaluated solutions, thus generating lower-quality offspring.

In summary, USEA effectively reduces computational expenses while maintaining the quality of new solutions, enhancing search efficiency.

\subsection{Runtime Comparison}
\label{sec:runtime_comparison}

\begin{figure}[ht!]
    \centering
    \includegraphics[width=0.9\linewidth]{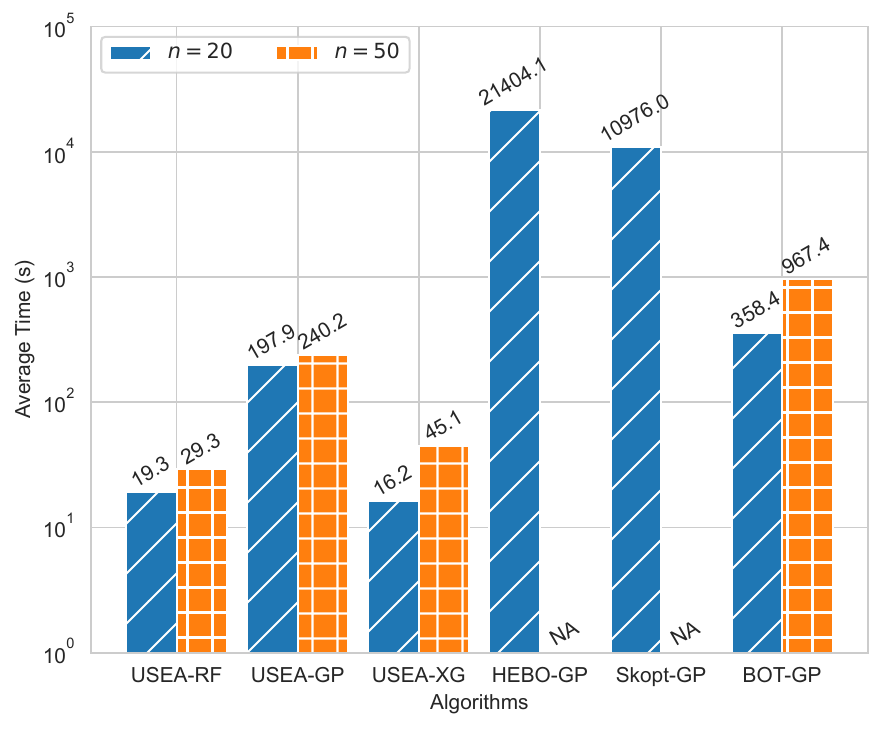}
    \caption{Runtime comparison of the USEA framework with other optimization algorithms on the LZG test suite with $n=20,50$. (NA signifies an impractical computational cost for experiment completion.)}
    \label{fig:runtime} 
    \end{figure}

Fig.~\ref{fig:runtime} presents the runtime statistics for three BOs and the USEA-EDA incorporating different surrogate models. The evaluations were uniformly conducted on the same hardware configuration (CPU: i9-13900k, Memory: 64GB, Ubuntu 22.04), addressing LZG01-04 problems at both 20 and 50 dimensions. This approach allowed for the determination of the average resolution times for each algorithm at different problem scales. The absence of data for HEBO-GP and Skopt-GP at $n=50$ signifies the computational cost was infeasible, hence it is denoted as `NA'.

The statistical results reveal that the UEDA algorithm significantly reduces runtime compared to the BO algorithms, particularly in comparison with HEBO-GP and Skopt-GP, by more than two orders of magnitude. This efficiency is attributed to UEDA's use of a more lightweight surrogate model and a reduced amount of model training data.

Fig.~\ref{fig:runtime} presents the runtime statistics for three BOs and the USEA utilizing different surrogate models. The evaluations were uniformly conducted on the same hardware configuration (CPU: i9-13900k, Memory: 64GB, Ubuntu 22.04), addressing LZG test suite at both 20 and 50 dimensions. This approach allowed for the determination of the average resolution times for each algorithm at different problem scales. The absence of data for HEBO-GP and Skopt-GP at \(n=50\) signifies that the computational cost was infeasible, hence denoted as `NA'.

The results indicate that USEA-EDA significantly reduces runtime compared to the BO algorithms, particularly against HEBO-GP and Skopt-GP, by more than two orders of magnitude. This efficiency is attributed to USEA's use of a more lightweight surrogate model and a reduced amount of model training data.

\subsection{Case Study between USEA and BO}

\begin{figure*}
    \centering
    \includegraphics[width=0.9\linewidth]{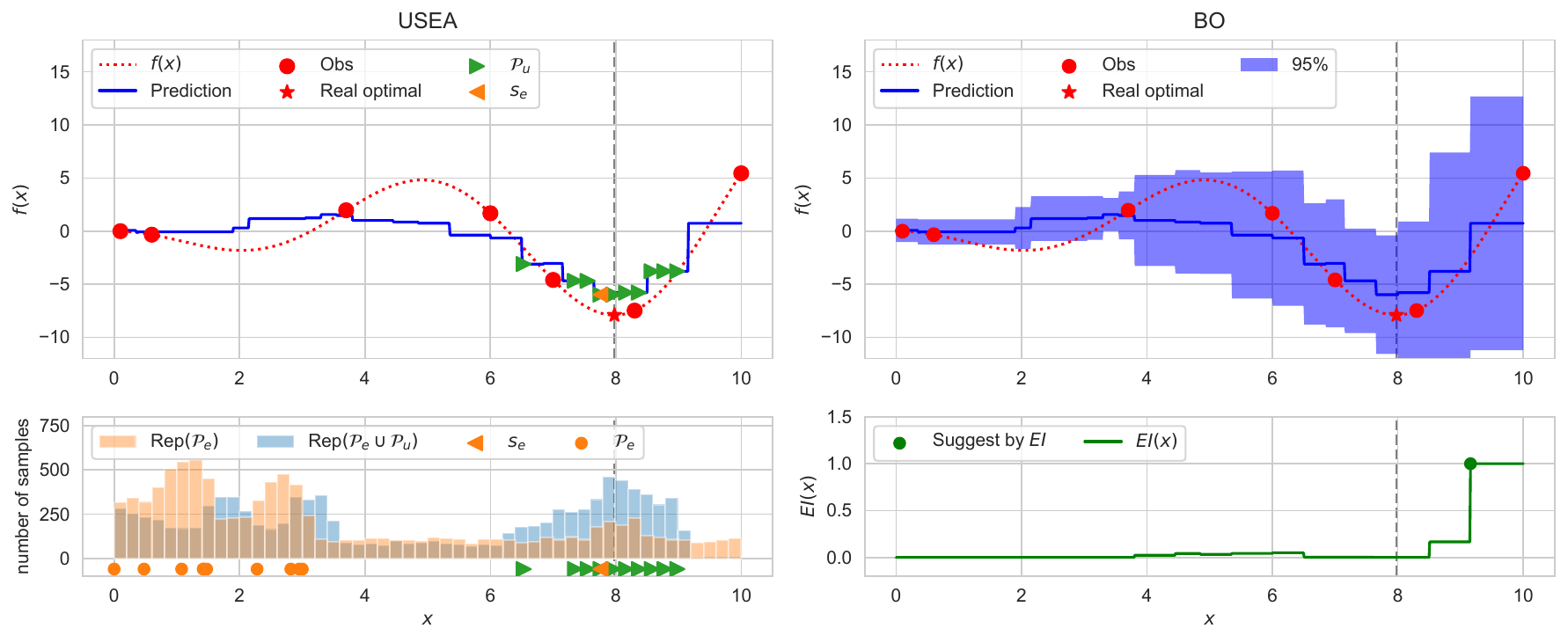}
    \caption{Comparison of USEA and BO in one-dimensional optimization. The top left shows the process of USEA selecting offspring; the bottom left shows the distribution of 10,000 offspring generated with or without using unevaluated solutions ($\mathcal{P}_u$), with the orange $\circ$ representing the assumed distribution of the previous generation population. The top right shows the mean and uncertainty of RF within the BO framework, and the bottom right shows the next sampling point calculated using the EI acquisition function.}
    \label{fig:bo_usea}
\end{figure*}

Based on the analyses in Section~\ref{sec:performance_evaluation} and Section~\ref{sec:runtime_comparison}, we can conclude that USEA has advantages over BO in terms of optimization performance and runtime. To explain this result, we use a 1-dimensional visualization case to highlight the runtime mechanism differences between the two algorithms. We select the case function used in Section~\ref{sec:motivation}. Fig.~\ref{fig:bo_usea} compares the USEA and BO algorithms in a one-dimensional optimization.

In Fig.~\ref{fig:bo_usea}, red dots represent the observed data used to train a random forest model. The blue line shows the model's predicted function. The right side of Fig.~\ref{fig:bo_usea} shows that the model's uncertainty estimation is inaccurate, particularly in the [8,10] interval where the estimated uncertainty is much higher than the actual prediction error. In such cases, using the current RF to compute the expected improvement for determining the next sampling point results in selections far from the optimal region (red star) and does not effectively reduce model uncertainty.

Conversely, USEA leverages the population-based search advantage of EAs. Despite the less accurate RF model, it selects points (green \(\triangleright\)) to diversify the population distribution~(orange $\circ $), generating more solutions near the optimal region and thus finding the optimal solution more quickly. This simple case illustrates how USEA, without additional evaluation costs, uses un-evaluated solutions to mitigate model error and increase the probability of generating offspring in the optimal region. In contrast, BO relies on sequential optimization and selection based on acquisition functions, which can be misled by model inaccuracies and uncertainty estimation errors, reducing optimization efficiency. Further research is needed to explore the deeper underlying reasons and differences in higher dimensions and iterative contexts.

\section{Conclusion}\label{sec:conclusion}

In this paper, we propose a Un-evaluation solution assisted evolutionary algorithm~(USEA) for expensive optimization problems. The basic idea behind USEA is that in solving EOPs using SAEA, the reduced disparity between adjacent parent populations leads to reproduction operators failing to generate sufficiently high-quality offspring, thereby affecting the algorithm's search efficiency. On the other hand, some high-quality solutions evaluated by the surrogate model, while not meeting the requirements for real evaluation, are still of sufficient quality to alter the population distribution and drive the population towards the optimal regions. By innovatively integrating these observations, a general strategy emerged: using the model to select a number of high-quality solutions for generating new solutions without undergoing real evaluation. This concept is specifically implemented across three operators: GA, DE, and EDA. Experimental results demonstrate that the proposed strategy enhances the performance of all three operators. Moreover, algorithms incorporating this strategy exhibit strong competitiveness against mainstream SAEA and BO algorithms. Notably, even when using only a Random Forest model, the performance achieved matches that of BO algorithms employing Gaussian Processes, while requiring only 0.1\% of the computation time.

Future research directions include refining the strategies to enhance their effectiveness. For instance, more sophisticated strategies could be designed for updating the population ($\mathcal{P}_e$) and the model training data, selecting un-evaluated solutions based on their distance from the current population, and using non-regression-based surrogate models. Furthermore, this work focused on single-objective optimization problems, and extending the approach to multi-objective optimization problems warrants further investigation.

\bibliographystyle{elsarticle-num} 
\bibliography{bare_jrnl.bib}

\end{document}